\PassOptionsToPackage{%
  colorlinks=true,
  linkcolor=blue,
  citecolor=green,
  urlcolor=magenta
}{hyperref}

\documentclass[sigconf,screen,nonacm]{acmart}
\AtBeginDocument{%
  }

\usepackage[utf8]{inputenc} 
\usepackage[T1]{fontenc}    
\usepackage{xcolor}
\usepackage{url}            
\usepackage{booktabs}       
\usepackage{amsfonts}       
\usepackage{nicefrac}       

\usepackage[font=small]{caption}
\usepackage{filecontents}
\usepackage{color}

\usepackage{amsmath, amsthm, bm, mathtools}
\usepackage{microtype}

\usepackage{pifont}

\usepackage{algorithm}     
\usepackage{algorithmic}   

\usepackage{graphicx}
\usepackage{subfig}

\usepackage{multirow}
\usepackage[font=small]{caption}

\DeclareMathOperator*{\argmin}{arg\,min}

\begin{document}

\title{Spectral Survival Analysis}


\newcommand{\arxiv}[2]{#1} 

\newcommand{\dataset}{\mathcal{D}}
\newcommand{\loss}{\mathcal{L}}
\newcommand{\vpi}{\bm{\pi}}
\newcommand {\mx}{\bm{X}}
\newcommand {\mW}{\bm{W}}
\newcommand{\vx}{\bm{x}}
\newcommand{\vr}{\bm{r}}
\newcommand{\vW}{\bm{W}}
\newcommand{\vtheta}{\bm{\theta}}
\newcommand{\mtheta}{\bm{\theta}}
\newcommand{\mzeta}{\bm{\zeta}}
\newcommand{\vu}{\bm{u}}
\newcommand{\vone}{\bm{1}}
\newcommand{\vzero}{\bm{0}}
\newcommand{\vy}{\bm{y}}
\def\vz{{\bm{z}}}
\newcommand{\veta}{\bm{\eta}}
\newcommand {\obt}{O}
\newcommand {\et}{T}
\newcommand {\ct}{C}
\newcommand{\ind}{\bm{1}} 
\def\mA{{\bm{A}}}
\def\mB{{\bm{B}}}
\def\mC{{\bm{C}}}
\def\mD{{\bm{D}}}
\def\mE{{\bm{E}}}
\def\mF{{\bm{F}}}
\def\mG{{\bm{G}}}
\def\mH{{\bm{H}}}
\def\mI{{\bm{I}}}
\def\mJ{{\bm{J}}}
\def\mK{{\bm{K}}}
\def\mL{{\bm{L}}}
\def\mM{{\bm{M}}}
\def\mN{{\bm{N}}}
\def\mO{{\bm{O}}}
\def\mP{{\bm{P}}}
\def\mQ{{\bm{Q}}}
\def\mR{{\bm{R}}}
\def\mS{{\bm{S}}}
\def\mT{{\bm{T}}}
\def\mU{{\bm{U}}}
\def\mV{{\bm{V}}}
\def\mW{{\bm{W}}}
\def\mX{{\bm{X}}}
\def\mY{{\bm{Y}}}
\def\mZ{{\bm{Z}}}
\def\mPhi{{\bm{\Phi}}}
\def\mLambda{{\bm{\Lambda}}}
\def\mSigma{{\bm{\Sigma}}}
\newcommand{\E}{\mathbb{E}}
\newcommand{\PL}{\mathcal{L}_{\mathtt{P}}}
\newcommand{\NLL}{\mathcal{L}_{\mathtt{NLL}}}
\newcommand{\Ls}{\mathcal{L}}
\newcommand{\R}{\mathbb{R}}
\newcommand{\emp}{\tilde{p}}
\newcommand{\lr}{\alpha}
\newcommand{\reg}{\lambda}
\newcommand{\rect}{\mathrm{rectifier}}
\newcommand{\softmax}{\mathrm{softmax}}
\newcommand{\sigmoid}{\sigma}
\newcommand{\softplus}{\zeta}
\newcommand{\KL}{D_{\mathrm{KL}}}
\newcommand{\Var}{\mathrm{Var}}
\newcommand{\standarderror}{\mathrm{SE}}
\newcommand{\Cov}{\mathrm{Cov}}
\newcommand{\normlzero}{L^0}
\newcommand{\normlone}{L^1}
\newcommand{\normltwo}{L^2}
\newcommand{\normlp}{L^p}
\newcommand{\normmax}{L^\infty}

\newtheorem{thm}{Theorem}
\newtheorem{lem}{Lemma}
\newtheorem{cor}{Corollary}




\author{Chengzhi Shi}
\affiliation{%
 \institution{Northeastern University}
 \city{Boston}
 \state{MA}
 \country{USA}}
\email{shi.cheng@northeastern.edu}

\author{Stratis Ioannidis}
\affiliation{%
 \institution{Northeastern University}
 \city{Boston}
 \state{MA}
 \country{USA}}
\email{ioannidis@northeastern.edu}





\begin{abstract}
  Survival analysis is widely deployed in a diverse set of fields, including healthcare, business, ecology, etc. The Cox Proportional Hazard (CoxPH) model is a semi-parametric model often encountered in the literature. Despite its popularity, wide deployment, and numerous variants, scaling CoxPH to large datasets and deep architectures poses a challenge, especially in the high-dimensional regime.  We identify a fundamental connection between rank regression and the CoxPH model: this allows us to adapt and extend the so-called spectral method for rank regression to survival analysis. Our approach is versatile, naturally generalizing to several CoxPH variants, including deep models. We empirically verify our method's scalability on multiple real-world high-dimensional datasets; our method outperforms legacy methods w.r.t. predictive performance and efficiency.

\end{abstract}




\maketitle

\section{Introduction}
\label{sec: intro}
The goal of survival analysis is to regress the distribution of event times from external covariates. Survival analysis is widely deployed in a diverse set of fields, including healthcare \cite{murtaugh1994primary, Ganssauge2017, Zupan1999}, business~\cite{FADER200776}, economics~\cite{danacica2010using}, and ecology~\cite{lebreton1993statistical}, to name a few. A prototypical example in healthcare is predicting the time until a significant medical event (e.g., the onset of a disease, the emergence of a symptom, etc.) from a patient's medical records \cite{Ganssauge2017, Zupan1999}. Another is regressing counting process arrival rates from, e.g., past events or other temporal features \cite{AalenBook}; this has found numerous applications in modeling online user behavior \cite{chen2023gateway,zhang2014multi, barbieri2016improving}.

Methods implementing regression in this setting are numerous and classic~\cite{CoxPH1972, EH2021, Wei1992TheAF}. Among them, the Cox Proportional Hazard (CoxPH) model is a popular semi-parametric model. The ubiquity of CoxPH is evidenced not only by its wide deployment and use in applications \cite{lane1986application, liang1990applicationepid, wong2011applicationchurn}, but also by its numerous extensions~\cite{subgroupreweight, hu2021subgroup},  including several recently proposed deep learning variants~\cite{Katzman2018DeepSurvPT, CoxTimeCC}.

Despite its widespread use, CoxPH and its variants lack scalability, especially in the high-dimensional regime. This scalability challenge comes from the partial likelihood that CoxPH optimizes \cite{CoxPH1972}, which {incorporates all samples in a single Siamese  objective~\citep{chicco2021siamese}. Previous works reduce the computational complexity of CoxPH variants through mini-batching \cite{CoxTimeCC, zhu2016deepconvsurv}; however, this leads to biased gradient estimation and reduces predictive performance in practice~\cite{whybatch}. Matters are only worse in the presence of deep models, which significantly increase the costs of training in terms of both computation and memory usage. As a result, all present deployments of methods with original data input are limited to shallow networks \cite{Katzman2018DeepSurvPT, coxnnet2018cox, CoxTimeCC, YAO2020101789, SurvCNN21, yin2022convolutional, zhu2016deepconvsurv} or require the use of significant subsampling and dimensionality reduction techniques~\cite{SurvCNN21,wang2018comprehensive} to be deployed in realistic settings.



\fussy

In this work, we leverage an inherent connection between CoxPH and \emph{rank regression} \cite{yildiz20a,yildiz21a,Burges2005LearningTR,Cao2007LearningTR,yildizTKDD}, i.e., regression of the relative order of samples from their features, using ranking datasets. Recently, a series of papers \cite{maystre2015fast,yildiz20a,yildiz21a,yildizTKDD} have proposed spectral methods to speed up ranking regression and to deal with memory requirements similar to the ones encountered in survival analysis. Our main contribution is to observe this connection and adapt these methods to the survival analysis domain. 
\begin{figure*}[t]  
    \centering
    \includegraphics[width=\linewidth]{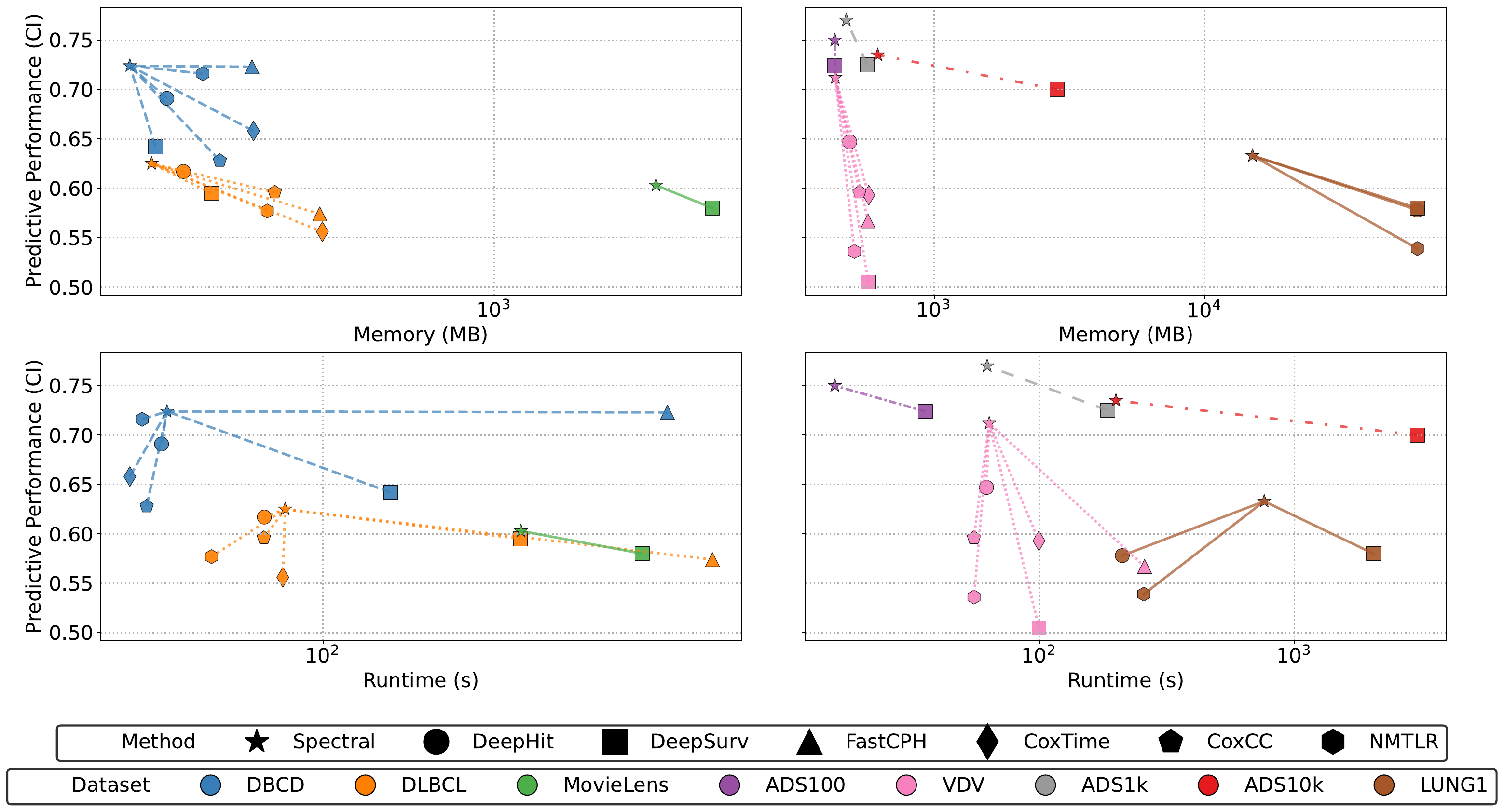}%
    \caption{Comparison of our proposed \textsc{Spectral} method against SOTA competitors on eight datasets, w.r.t.~predictive performance (CI, $\uparrow$), runtime (s, $\downarrow$), and memory (MB, $\downarrow$). All methods executed over the same dataset are connected to the proposed method \textsc{Spectral}; not all methods were applicable to all datasets, and some ran out of memory (see Table.~\ref{tab: LUNG1 performance.}). \textsc{Spectral} consistently achieves superior predictive performance over competitors while using less memory. It also exhibits marked acceleration over the \textsc{DeepSurv} base model, with which it shares the same objective; it is comparable in runtime to other methods with simpler objectives which, nevertheless, perform worse in predictive performance. }
    \label{fig:teaser_standard}
\end{figure*}

Overall, we make the following contributions: 
\begin{itemize}
    \item We identify a fundamental connection between rank regression and survival analysis via the CoxPH model and  variants. 
    \item Inspired by the spectral method in \cite{maystre2015fast} and \cite{yildiz21a}, we propose to use spectral method to achieve the staionary point solutions to CoxPH in the survival analysis setting. To the best of our knowledge, ours is the first spectral approach used to attack survival analysis regression. 
    \item By introducing the weight coefficients, we make this approach versatile: in particular, it naturally generalizes to several variants of CoxPH, including deep models, that are extensively used in applications of survival analysis. 
    \item We empirically verify the ability of our method to scale on multiple real-world high-dimensional datasets, including a high-dimensional CT scan dataset. We can scale deep CoxPH models such as DeepSurv~\cite{Katzman2018DeepSurvPT} and outperform state-of-the-art methods that rely on dimensionality reduction techniques, improving  predictive performance and memory consumption, while being better or comparable in runtime (see Fig.~\ref{fig:teaser_standard} and Sec.~\ref{subsec: results}).
\end{itemize}

The remainder of this paper is organized as follows. In Sec.~\ref{sec: related work}, we review the relevant literature and contextualize our contributions. Sec.~\ref{sec: preliminaries} provides the necessary preliminaries for understanding our approach. In Sec.~\ref{sec: method}, we introduce our spectral method for scaling CoxPH-based survival analysis. In Sec.~\ref{sec: extensions} we demonstrate how to extend this approach to other classical survival analysis models such as the Accelerated Failure Time (AFT) model \cite{Wei1992TheAF}. Finally, we empirically verify the predictive performance and scalability of our proposed method in Sec.~\ref{sec: experiments}.

\section{Related Work}
\label{sec: related work}

\sloppy \noindent \textbf{Survival Analysis.} Traditional statistical survival analysis methods are categorized as non-parametric, parametric, and semi-parametric~\cite{survivalcategory}. 
Ror completeness, we review non-parametric and parametric approaches in App.~\arxiv{\ref{app: related work}}{A~\cite{arxiv}}. In short, non-parametric methods, exemplified by the Kaplan-Meier \cite{Kaplan1992} and Nelson-Aalen estimators \cite{AalenBook}, estimate baseline hazard functions directly without making distributional assumptions. However, they do not explicitly capture the effects of features, limiting their ability to generalize out-of-sample. In contrast, parametric methods explicitly assume survival times follow parametrized distributions, such as the exponential \cite{WITTEN19927, jenkins2005survival}, Weibull~\cite{carroll2003use}, and log-normal distributions~\cite{Wei1992TheAF}. In turn, distribution parameters can naturally be regressed from features~\cite{AalenBook}. However, although parametric models offer simplicity and interpretability, they may have high bias and not fit the data well when the underlying distributional assumptions are violated.

\fussy Semi-parametric models, like  CoxPH~\cite{CoxPH1972}, incorporate features without assuming a specific form for the baseline hazard, striking a better bias-variance tradeoff.  Their popularity has led to numerous practical applications \cite{Katzman2018DeepSurvPT,mobadersany2018predicting, tran2021deep} as well as CoxPH deep variants from the machine learning community~\cite{zhu2016deepconvsurv, lee2018deephit,Deepsurvranking2019}. For example, DeepSurv \cite{Katzman2018DeepSurvPT} replaces the linear predictor in standard CoxPH with a fully connected neural network \cite{SELU17} for predicting relative hazard risk. DeepConvSurv~\cite{zhu2016deepconvsurv} combines CoxPH with a CNN structure, to regress parameters from image datasets. \textsc{CoxTime} \cite{CoxTimeCC} stacks time into features to make the risk model time-dependent. \textsc{DeepHit} \citep{lee2018deephit} discretizes time into intervals and introduces a ranking loss to classify data into bins on the intervals.
Nnet-survival \citep{NNet} parameterizes discrete hazard rates by NNs and the hazard function is turned into the cumulative product of conditional probabilities based on intervals.

Nevertheless,  the above methods either do not scale w.r.t. sample size and high dimensionality \cite{Katzman2018DeepSurvPT,zhu2016deepconvsurv,CoxTimeCC} or resort to mini-batch SGD ~\cite{NNet,lee2018deephit,CoxTimeCC}, which as discussed below reduces predictive performance. Scalability is hampered by training via gradient descent against the partial likelihood loss,  which as discussed in Sec.~\ref{sec: method} yields a Siamese network (SNN) penalty: this incurs significant computational and memory costs (see details in Sec.~\ref{subsec: results}). As a result, both model and dataset sizes considered by prior art \cite{Katzman2018DeepSurvPT,zhu2016deepconvsurv,CoxTimeCC} are small: For example, \textsc{DeepSurv}  only adopts a shallow two-layer Perceptron and the datasets contain hundreds of samples. 
Approaches to resolve this include introducing mini-batch SGD (as done by Nnet-survival~\cite{NNet}, DeepHit~\cite{lee2018deephit}, and CoxCC~\cite{CoxTimeCC}) or decreasing input size by, e.g., handling patches of images rather than the whole image \cite{aerts2014decoding, Haar2019, Brag2023}. Both come with a predictive performance degradation. We demonstrate this extensively in our experiment section (see Table.~\ref{tab: LUNG1 performance.}). From a theoretical standpoint, in contrast to SGD over traditional decomposible ML objectives, mini-batch SGD w.r.t.~partial likelihood introduces bias in gradient estimation (see App.~\arxiv{~\ref{app: biased est}}{C~\cite{arxiv}}). 
Alternative approaches like SODEN \cite{tang2022soden}  maximize the full (rather than the partial) likelihood, for which SGD is unbiased, this comes with the drawbacks of using the full likelihood (see also App.~\arxiv{\ref{app: related work}}{A} \cite{arxiv}); moreover, the computational cost of SODEN remains high in practice, as reported by ~\citet{wu2023neural}. 

\noindent \textbf{Rank Regression.}
In ranking regression problems \cite{guo19a, yildiz21a, yildizTKDD, RankSVM}, the goal is to regress a ranking function from sample features. RankSVM \cite{RankSVM} learns to rank via a linear support vector machine. RankRLS \cite{RankRLS} regresses rankings with a regularized least-square ranking cost function based on a preference graph. Another line of work regresses from pairwise comparisons \citep{lee2023towards} by executing maximum-likelihood estimation (MLE) based on the Bradley-Terry model~\cite{bradley1952rank}, in either a shallow~\cite{tian2019severity, guo2018experimental} or deep~\cite{doughty2018s} setting. Several other papers~\cite{han2018dateline, doughty2018s} regress from partial rankings via the Placket-Luce model \cite{PlackettLuce1975}, which generalizes Bradley-Terry. All the above methods apply the traditional Newton method to maximize the likelihood function, corresponding to training via siamese network penalty. This is both memory and computation-intensive. Yildiz et al \cite{yildiz20a, yildiz21a} accelerate computations and reduce the memory footprint by applying a spectral method proposed by Maystre and Grossglauser~\cite{maystre2015fast} to the regression setting; we review both in App.~\arxiv{\ref{app: spectral}}{B~\cite{arxiv}}. From a technical standpoint, we extend the analysis of \citet{yildiz21a} by incorporating censorship and weights in the loss, and showing that its stationary points can still be expressed as solutions of a continuous Markov Chain. This extension is crucial in tackling a broad array of CoxPH variants: we show how to address the latter via a novel alternating optimization technique, combining spectral regression with a Breslow estimate of baseline hazard rates.

Leveraging the connection between the Placket-Luce model in ranking regression and the partial likelihood of the Cox model in survival analysis, we transfer the spectral method of Yildiz et al. to reduce memory costs and scale the Cox Proportional Hazard model. To account for (a) censoring and (b) the baseline hazard rate, we extend Yildiz et al. to a framework that works with both censorship in survival analysis and introduce weights to enable the extension to other survival analysis methods inlcuding CoxPH, its extensions, and several other survival analysis models such as the Accelerated Failure Model (AFT) scalable via the \textsc{Spectral} method (see App.~\arxiv{\ref{app: extensions}}{D~\cite{arxiv}}).

\begin{figure}[t!]  
    \centering
    \includegraphics{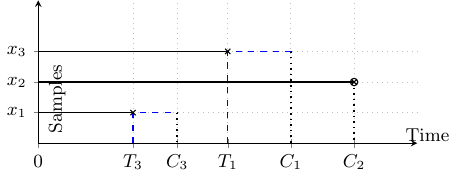}
        
    \caption{Illustration of the standard survival analysis problem setting. Each sample is associated with a $d$-dimensional feature vector $\vx_i \in \mathbb{R}^d$, an {event time} $\et_i>0$, and a {censoring time} $\ct_i>0$. Event times are observed only if they occur before the censoring time: for example, no event is observed for $x_2$ before the censoring time $C_2$.}
  \label{fig: illustration}
\end{figure}

\section{Preliminaries}
\label{sec: preliminaries}

We provide a short review of survival analysis fundamentals; the subject is classic: we refer the interested reader to Aalen et al \cite{AalenBook} for a more thorough exposition on the subject, as well as to App.~\arxiv{\ref{app: related work}}{A~\cite{arxiv}}.

\noindent\textbf{Survival Analysis Problem Setup.}
Consider a dataset $\dataset$ comprising $n$ samples, each associated with a $d$-dimensional feature vector $\vx_i \in \R^d$. Each sample is additionally associated with an \emph{event time} $\et_i>0$ and a \emph{censoring time} $\ct_i>0$ (see Fig.~\ref{fig: illustration}). For example, each sample could correspond to a patient, the event time may denote a significant event in the patient's history (e.g., the patient is cured, exhibits a symptom, passes away, etc.), while the censoring time captures the length of the observation period:  events happening \emph{after} the censoring time are not observed.

As shown in Fig~\ref{fig: illustration}, because we are not able to observe events outside the observation period, the dataset $\mathcal{D}$ is formally defined as $\mathcal{D}=\{\vx_i,\obt_i,\Delta_i\}^n_{i=1},\label{eq:dataset}$ where $O_i=\min(T_i,C_i)$ is the \emph{observation time} (either event or censoring) and $\Delta_i\equiv \ind_{\et_i\leq \ct_i}$ is the \emph{event indicator}, specifying whether the event was indeed observed. Note that $O_j=T_j$, i.e., an event time is observed, iff $\Delta_i=1$. The goal of survival analysis is to regress the distribution of the (only partially observed) times $\et_i$ from feature vectors $\vx_i$, using dataset $\mathcal{D}$. The act of censoring, i.e., the fact that events happening outside the observation period are not observed, distinguishes survival analysis from other forms of regression.

Typically, we express the distribution of event times via the so-called \emph{survival} and \emph{hazard} functions. The survival function is the complementary cumulative distribution function, i.e.,
$S(t|\vx)=P(T>t|\vx)\label{eq:surv},$ i.e., it describes the likelihood that the event occurs after $t$. Given a sample $\vx$, the hazard function is:
\begin{align}\label{eq: hazard}
    \lambda(t|\vx)=\lim_{\delta \to 0} \frac{P(t\leq\! T\!\leq t\!+\!\delta|T\geq t,\vx)}{\delta}=\frac{f(t|\vx)}{S(t|\vx)}=-\frac{S'(t|\vx)}{S(t|\vx)},
\end{align}
where $f(t|\vx)=-S'(t|\vx)$ is the density function of $\et$ given $\vx$. Intuitively,
the hazard function captures the rate at which events occur, defined via the probability that an event happens in the infinitesimal interval $(t,t+\delta]$.  
Eq.~\eqref{eq: hazard} implies that we can express the survival function via the cumulative hazard via
 $   S(t|\vx)=\exp{\left(-\int_0^t \lambda(s|\vx)\, ds\right)}.$


\noindent\textbf{The CoxPH Model.} \label{subsec: hazard models} One possible approach to learning feature-dependent hazard rates is to assume that event times follow a well-known distribution (e.g., exponential, Weibull, log-normal, etc.), and subsequently regress the parameters of this distribution from features using maximum likelihood estimation ~\citep{AalenBook}. However, as discussed in App.~\arxiv{\ref{app: related work}}{A~\cite{arxiv}}, the distribution selected by such a parametric approach may not fit the data well, introducing estimation bias. The Cox Proportional Hazard (CoxPH)~\citep{CoxPH1972} is a popular semi-parametric method that directly addresses this issue.
%
Formally, define $[n]\coloneqq \{1,\dots, n\}$. Then, CoxPH assumes that the hazard rate has the form:
\begin{align}
    \lambda(t|\vx_i)=\lambda_0(t)e^{\mtheta^T \vx_i}, \quad i\in [n]\equiv\{1,\ldots,n\},\label{eq:coxhazard}
\end{align}
where $\lambda_0(\cdot)$ is a common baseline hazard function across all samples, and $e^{\mtheta^T \vx_i}$ is the relative risk with $\mtheta\in \mathbb{R}^d$ being a parameter vector. 

Parameter vector $\mtheta$ is learned by maximizing the so-called \emph{partial likelihood}:
\allowdisplaybreaks
\begin{align}
\mathcal{L}_{\mathtt{P}}(\dataset|\mtheta)&=\prod_{i=1}^n\left[\frac{\lambda(T_i|\vx_i)}{\sum_{j\in \mathcal{R}(T_i)}\lambda(T_i|\vx_j)}\right]^{\Delta_i}\label{eq: CoxPH plik0}\\
&\displaybreak[0]\stackrel{\text{Eq.}~\eqref{eq:coxhazard}}{=}\prod_{i=1}^n\left[   \frac{ e^{\mtheta^\top \vx_i}}{\sum_{j\in \mathcal{R}(T_i)}  e^{\mtheta^\top \vx_j}}\right]^{\Delta_i}, \label{eq: CoxPH plik}
\end{align}
where $\mathcal{R}(t)\equiv \{j\in[n]: t<O_j\}$ be the set of at-risk samples at time $t$. Intuitively, this is called a partial likelihood because it characterizes the \emph{order} in which observed events occur, rather than the event times themselves. Indeed, the probability of the observed order of events is given by \eqref{eq: CoxPH plik}: this is because the probability that sample $i$ experiences the event before other at-risk samples at time $t$ are proportional to the hazard $\lambda(t|\vx_i)$. Crucially, as the hazard rate is given by Eq.~\eqref{eq:coxhazard}, Eq.~\eqref{eq: CoxPH plik0} simplifies to Eq.~\eqref{eq: CoxPH plik}, that \emph{does not involve $\lambda_0$}. Hence, vector $\mtheta$ can be directly regressed from data \emph{separately} from $\lambda_0$, by minimizing the negative log-likelihood:
\begin{align}
\hat{\mtheta} 
=\argmin_{\mtheta} \sum_{i=1}^n\Delta_i \left[\log \sum_{j\in \mathcal{R}(T_i)} e^{\mtheta^\top \vx_j}-\mtheta^\top \vx_i\right].\label{eq:nll}
\end{align}
This is typically obtained through standard techniques, e.g., gradient descent or Newton's method.
Having learned the parameter vector, the next step is to estimate the baseline hazard function. This can in general be done by an appropriately modified non-parametric method, such as Nelson-Aalen \cite{NelsonAalen1978} or Kaplan-Meier \cite{Kaplan1992}. For example, the lifelines package~\cite{davidson2019lifelines}  uses the Breslow estimator \cite{davidson2019lifelines, Xia2018Breslow}  to obtain the baseline hazard rate 
$\hat{\lambda}_0(t)= 1/\sum_{i\in \mathcal{R}(t)} h_{\mtheta}(\vx_i)$, with $h_{\mtheta}(\vx) = e^{\hat{\mtheta}^\top \vx_i}$. We review additional non-parametric methods in App.~\arxiv{\ref{app: related work}}{A~\cite{arxiv}}.

\noindent\textbf{\textsc{DeepSurv}.}\label{sec:deepsurv}
To learn more complicated representations, Jared et al. \cite{Katzman2018DeepSurvPT} replace the linear model in CoxPH with a neural network $h_{\mtheta}\colon \R^d \to \R$ with parameters $\mtheta\in \R^p$, leading to  a hazard rate of the form:
\begin{align}\label{eq:deepsurvhaz}
\lambda(t|\vx_i)=\lambda_0(t)e^{h_{\mtheta}( \vx_i)}, \quad i\in [n]. 
\end{align}
Parameters $\mtheta$ are again learned by maximizing the  partial likelihood as in Eq.~\eqref{eq: CoxPH plik}, by performing, e.g.,  gradient descent on 
the corresponding negative log-likelihood loss
\begin{align}\label{eq: DeepSurv  Loss}
     \NLL(\dataset,\mtheta)=-\sum_{i=1}^n\Delta_i \left[ h_{\mtheta}(\vx_i)-\log \sum_{j\in \mathcal{R}_i} e^{h_{\mtheta}(\vx_j)}\right]
\end{align}
Finally, the baseline hazard function $\lambda_0(\cdot)$ can again be estimated by the corresponding modified Nelson-Aalen estimator \cite{NelsonAalen1978},  replacing the linear model with $h_{\mtheta}(\cdot)$. By setting weights to $1$ and replacing the linear predictor with the neural network, we can directly apply the theorem to \textsc{DeepSurv}.

\noindent\textbf{Advantages of \textsc{CoxPH}/\textsc{DeepSurv}}. The above end-to-end procedures have several benefits. Being semi-parametric allows regressing the hazard rate from features $\vx$ while maintaining the flexibility of fitting $\lambda_0$ to data through the Breslow estimate:  this significantly reduces bias compared to committing to, e.g., a log-normal distribution. Moreover, the parametrization via Eq.~\eqref{eq:coxhazard} naturally generalizes to several extensions (see App.~\arxiv{\ref{app: extensions}}{D~\cite{arxiv}}). Finally, as discussed next, a solution to  Eq.~\eqref{eq:nll} is obtainable via a spectral method~\cite{maystre2015fast,yildiz20a,yildiz21a}; establishing this is one of our main contributions.

\section{Methodology}
\label{sec: method}
As discussed in Sec.~\ref{sec: preliminaries}, CoxPH optimizes the partial likelihood (Eq.~\eqref{eq: CoxPH plik}). However, this likelihood \emph{incorporates all samples in a single batch/term in the objective}. This is exemplified by Eqs.~\eqref{eq:nll} and \eqref{eq: DeepSurv  Loss}: 
all samples $j\in\mathcal{R} (T_i)$ need to be included in the loss when computing the term corresponding to sample $i$.  This is exacerbated in the case of \textsc{DeepSurv}  (Eq.~\eqref{eq: DeepSurv  Loss}), as it requires loading both the samples and $|\mathcal{R}(T_i)|$ identical copies of the neural network in the penalty to compute the loss.\footnote{This structure is also referred to as a Siamese network loss~\citep{chicco2021siamese} in the literature.} This makes backpropagation extremely expensive in terms of both computation and memory usage. In turn, this significantly limits the model architectures as well as the number and dimensions of the datasets that can be combined with \textsc{DeepSurv}  and the other extensions mentioned above. As a result, all present deployments of these methods are limited to shallow networks \cite{Katzman2018DeepSurvPT, coxnnet2018cox, CoxTimeCC, YAO2020101789, SurvCNN21, yin2022convolutional, zhu2016deepconvsurv} or require the use of subsampling and dimensionality reduction techniques~\cite{SurvCNN21,wang2018comprehensive} such as extracting patches and slices, which inevitably lead to information loss.  

We directly address this issue in our work. In particular, we show that spectral methods~\cite{maystre2015fast, yildiz20a,yildiz21a} can be used to decouple the minimization CoxPH and DeepSurv losses from fitting the neural network. Thus, we convert the survival analysis problem (likelihood optimization) to a regression problem, fitting the neural network to intrinsic scores that minimize the loss. Crucially, these scores can be efficiently computed via a spectral method, while fitting can be done efficiently via stochastic gradient descent on \emph{a single sample/score} at a time. As we will show in Sec.~\ref{sec: experiments}, this yields both time and memory performance dividends, while also improving the trained model's predictive performance.  By leveraging Theorem~\ref{thm: ss} we can extend the spectral method to scale other models beyond DeepSurv, such as AFT \cite{chen2003accelerated}, and Heterogeneous CoxPH~\cite{hu2021subgroup}, as well as to regress the arrival rate of a counting process. We focus in this section on CoxPH/DeepSurv for brevity, and present such extensions in detail in App.~\arxiv{\ref{app: extensions}}{D~\cite{arxiv}}.




\subsection{Regressing Weighted CoxPH and DeepSurv}
\label{subsec: weighted DeepSurv}
We describe here how to apply the spectral methods introduced by \citet{maystre2015fast} and  \citet{yildiz21a} to the survival analysis setting; we first present this here in the context of the so-called weighted CoxPH (DeepSurv) model and but note that it can be extended to the array of hazard models presented in App.~\arxiv{\ref{app: extensions}}{D~\cite{arxiv}}. Our main technical departure is in handling heterogeneity in the loss, as induced by weights, which \citet{yildiz21a} do not consider. This is crucial in generalizing our approach to CoxPH extensions: this also technically involved, and incorporatesalternating between regression and  Breslow estimate of the baseline hazard (see App.~\arxiv{\ref{app: extensions}}{D~\cite{arxiv}}). 

\noindent{\bf Reduction to a Spectral Method}.
We apply the \textsc{Spectral} approach by \cite{yildiz21a} to minimize the following general negative-log likelihood:
\begin{align}\label{eq: reweight DeepSurv nll}
   \NLL(\dataset,\mtheta) =  \frac{1}{n}\!\sum_{i=1}^n \!{\Delta_i}\Big(\log\!\!\sum_{j\in \mathcal{R}_i}\!\!\mW_{ji} \tilde{h}_{\mtheta}(\vx_j)- \log  \mW_{ii} \tilde{h}_{\mtheta}(\vx_i) \Big).\!\!
\end{align}
where $\mW_{ji}\in\R_+$ denotes the weight for sample $j$ at time $T_i$ and we define $\mathcal{}{R}_i \equiv \mathcal{R}(T_i)$ for simplicity. Setting $ \tilde{h}_{\mtheta}(\vx)\equiv e^{\mtheta^\top \vx}$, we obtain the  Weighted CoxPH model;  setting it to be $ \tilde{h}_{\mtheta}(\vx)\equiv e^{\tilde{h}_{\mtheta}(\vx)}$, where ${h}_{\mtheta}(\vx)$ is a deep neural network we obtain a weighted version of \textsc{DeepSurv} ; finally, setting the weights to be one yields the standard CoxPH/DeepSurv models. Parameters in all these combinations can be learned through our approach.
\subsection{A Spectral Method}
Our goal is to minimize this loss. Following \cite{yildiz21a}, we reformulate this minimization as the following constrained optimization problem  w.r.t. $\vpi$, $\mtheta$:
 \begin{subequations}\label{eq: DeepSurv  problem}
      \begin{align}
     \text{Minimize}:& \sum_{i=1}^n {\Delta_i} \Big(\log\sum_{j\in \mathcal{R}_i}\mW_{ji} \vpi_j- \log  \mW_{ii} \vpi_i \Big)\\
     \text{subj.~to}: & \quad \vpi =  \tilde{h}_{\mtheta}(\mX), \vpi\geq 0,
 \end{align}
  \end{subequations}
where $\mX\coloneqq [\vx_1,\dots, \vx_n]^\top\in \R^{n\times d}$, $\tilde{h}_{\mtheta}\in \R^n_+$ is a map from this matrix to the images $ \tilde{h}_{\mtheta}(\vx_i)$ of every sample $\vx_i$,  $\vpi=[\vpi_1, \cdots, \vpi_n]^n \in \R_+^n$, and auxiliary variables $\vpi_i\in\R_+$, $i\in [n]$, are \emph{intrinsic scores} per sample: they are proportional to the probability that a sample $i$ experiences an event before other at-risk samples. 
We solve this equivalent problem via the Alternating Directions Method of Multipliers (ADMM) \cite{boyd2011admm}. In particular, 
we define the augmented Lagrangian of this problem as:
\begin{align}\label{eq:weighted DeepSurv lagrangian}
\begin{split}
    \loss(\vpi,\mtheta,\vu)\equiv & \sum_{i=1}^n {\Delta_i}\left(\log\sum_{j\in \mathcal{R}_i}\mW_{ji} \vpi_j- \log \mW_{ii} \vpi_i  \right)\\
    &+\vu^\top (\vpi - \tilde{h}_{\mtheta}(\mx))+\rho \mathcal{D}_{KL}(\vpi||h_{\mtheta}(\mx)),
    \end{split}
\end{align}
where $\mathcal{D}_{KL}(\cdot||\cdot)$ is the KL-divergence. Subsequently, ADMM proceeds iteratively, solving the optimization problem in Eq.~\eqref{eq:weighted DeepSurv lagrangian} in an alternating fashion, optimizing along each parameter separately:
\begin{subequations}
\label{eq:admm}
\begin{align}
\vpi^{(k+1)}=&\argmin_{\vpi\geq 0} \loss(\vpi,\mtheta^{(k)},\vu^{(k)}) \label{lin pi}\\
    \mtheta^{(k+1)}=&\argmin_{\mtheta} \loss(\vpi^{(k+1)},\mtheta,\vu^{(k)}) \label{lin theta}\\
    \vu^{(k+1)}=& \vu^{(k)} +\rho (\vpi^{(k+1)} - h_{\mtheta^{(k+1)}}(\mx)). \label{dual params}
\end{align}  
\end{subequations}
This approach comes with several significant advantages. First, Eq.~\eqref{lin pi}, which involves the computation of the intrinsic scores, can be performed efficiently via a \textsc{Spectral} method. Second, Eq.~\eqref{lin theta} reduces to standard regression of the model w.r.t. the intrinsic scores, which can easily be done via SGD, thereby decoupling the determination of the intrinsic scores (that involves the negative log-likelihood) from training the neural network. Finally, Eq.~\eqref{dual params} is a simple matrix addition, which is also highly efficient. We describe each of these steps below.

\noindent\textbf{Updating the Intrinsic Scores (Eq.~\eqref{lin pi}).} We show here how the intrinsic score computation can be accomplished via a \textsc{Spectral} method. To show this, we first prove that the solution to ~\eqref{lin pi} can be expressed as the steady state distribution of a certain Markov Chain (MC): 
\begin{thm}\label{thm: ss}
The stationary point of equation~\eqref{lin pi} satisfies the balance equations of a continuous-time Markov Chain 
 with transition rates 
 \begin{subequations}
 \label{eq: P}
\begin{align}
\mP_{ji}(\vpi)&= \mu_{ji}+ \Delta_{ji}(\vpi),~\text{for}\\
\Delta_{ji}(\vpi) &=
\begin{cases}
\frac{2\vpi_i \sigma_i(\vpi) \sigma_j(\vpi)}{\sum_{t\in [n]_-}\vpi_t \sigma_t(\vpi)-\sum_{t\in [n]_+}\vpi_t \sigma_t(\vpi)} & \textit{if } j \in [n]_+\\&\text{and}~i\in [n]_- \\
0 & \textit{o.w.},\\
\end{cases}
\end{align}
\end{subequations}
where  $[n]_{+}=\{i:\sigma_i(\vpi)\geq 0\}$, $[n]_{-}=\{i:\sigma_i(\vpi)\leq 0\}$, and
\begin{subequations}
\label{eq: sigma and mu}
\begin{align}
\sigma_i(\vpi)&= \rho \frac{\partial D_p(\vpi||\tilde{h}_{\mtheta}(\mx))}{\partial \vpi_i}+u^{(k)}_i,\\ \mu_{ji}&=\sum_{\ell\in W_i\cap L_j}\Delta_\ell \left(\frac{\vW_{j,\ell}}{\sum_{t\in R_\ell} \vW_{t,i} \vpi_t}\right), \\
W_i&=\{\ell|i\in  R_\ell,  i=\argmin_{j\in R_\ell} O_j\},\\ L_i&=\{\ell|i\in R_{\ell},i \neq \argmin_{j\in R_\ell} O_j\}.
%
\end{align}
\end{subequations}
\end{thm}
We prove this in App.~\arxiv{\ref{app:proof}}{E~\cite{arxiv}}. The key technical contribution is the presence of weights, which enables the extension of the framework beyond CoxPH (see Sec.~\ref{sec: extensions} for more details).
Thus, we can obtain a stationary point solution to \eqref{lin pi} via Theorem~\ref{thm: ss}. This is done by applying the following iterations:
\begin{align}
    \vpi^{(k+1)}=\mathrm{ssd}(\mP(\vpi^{(k)})),\label{eq:ssd}
\end{align}
where $\mP(\vpi^{(k)})$ is the transition matrix given by Theorem~\ref{thm: ss}, and $\mathrm{ssd}(\cdot)$ is an operator that returns the steady state distribution of the corresponding continuous-time MC.  This can be computed, e.g., using the power method \cite{Kuczynski1992power}.


\noindent\textbf{Fitting the Model (Eq.~\eqref{lin theta}).}  In App.~\arxiv{\ref{app:max entropy}}{F~\cite{arxiv}}, we show that Eq.~\eqref{lin theta} is equivalent to the following problem:
\begin{displaymath}
    \min_{\mtheta} \sum_{i=1}^n\left( - \vu_i^\top h_{\mtheta}(\vx_i)+\rho \vpi_i^{(k+1)}\log (\tilde{h}_{\mtheta}(\vx_i))\right).
\end{displaymath}

%
This is a standard max-entropy loss minimization problem, along with a linear term, and can be optimized via standard stochastic gradient descent (SGD) over the samples.

\noindent\textbf{Updating the Dual Variables (Eq.~\eqref{dual params}).}
The dual parameters are updated by adding the difference between intrinsic scores and the prediction of the updated network.


\begin{algorithm}[t]
\caption{Spectral Weighted Survival Analysis}
\label{algo: ADMM}
\begin{algorithmic}[1]
\STATE \texttt{{\bfseries Procedure:} \hspace{1em}  SpectralWeightedSA($\vpi, \mtheta, \vu, \mW, \dataset$)}
\STATE Initialize $\hat{\vpi} \gets \frac{1}{n} \vone$; $\vu \gets \vzero$
\REPEAT
    \STATE $\vpi \gets$ IterativeSpectralRanking$(\rho, \vpi, \vu, \hat{\vpi}, \mW)$ \textit{// via Thm~\ref{thm: ss}}
    \STATE $\mtheta \gets$ SGD over Eq.~\eqref{lin theta}
    \STATE $\hat{\vpi} \gets \tilde{h}_{\mtheta}(\mx)$
    \STATE $\vu \gets \vu + \rho(\vpi - \hat{\vpi})$
\UNTIL{convergence}
\STATE \textbf{Return:} $\vpi, \mtheta$
\STATE ~
\STATE \texttt{{\bfseries Procedure:} \hspace{1em}  IterativeSpectralRanking($\rho, \vpi, \vu, \hat{\vpi}, \mW$)}
\REPEAT
    \STATE Compute $[\sigma_i(\vpi)]_{i=1}^n$ using Eq.~\eqref{eq: sigma and mu}
    \STATE Compute transition matrix $P(\vpi)$ via Eq.~\eqref{eq: P}
    \STATE $\vpi \gets$ steady-state distribution of $P(\vpi)$
\UNTIL{convergence}
\STATE \textbf{Return:} $\vpi$
\end{algorithmic}
\end{algorithm}

\subsection{Algorithm Overview and Complexity.} 
\textsc{Spectral} is summarized in Algorithm~\ref{algo: ADMM}.
We initialize $\vpi = \frac{1}{n} \vone \in \R^n$, $ \vu = \vzero \in \R^n$ and sample $\mtheta$ from a uniform distribution. Then, we iteratively: (1) optimize the scores $\vpi$ via Eq.~\eqref{lin pi} with Theorem~\ref{thm: ss} and power method; (2) optimize the parameters $\mtheta$ via Eq.~\eqref{lin theta} with GD; (3) update the dual parameter $\vu$ via Eq.~\eqref{dual params} until the scores converge. 

Before discussing the complexity of Algorithm~\ref{algo: ADMM}, we briefly review the complexity of DeepSurv. Let $n$ be the number of samples, $B$ a batch-size parameter, and $P$ the number of parameters of the neural network; note that, typically, $P\gg d$. DeepSurv variants rely on  siamese neural networks to compute the negative log partial likelihood loss (see Eq.~\eqref{eq: reweight DeepSurv nll}). This involves a summation of $n$ terms, each further comprising $O(n)$ terms (as $|\mathcal{R}_i|$ is $O(n)$). Put differently, even with SGD, each batch of size $B$ loads features of $O(n)$ samples and copies the neural network $O(n)$ times (once for every term in $\mathcal{R}_i$). Overall, the time complexity of an epoch is $O(Pn)$ with the ``ascending order'' trick \citep{simon2011ascend}, where $P$ is the number of weights in the neural network, and the memory complexity is $O(PnB)$. 

On the other hand, in each epoch, Alg.~\ref{algo: ADMM} (a) computes the intrinsic scores first via the power method, which is $O(n^2)$ and does not load model weights or sample features in memory, and (b) trains with a single neural network with complexity $O(Pn)$ with a standard cross-entropy loss. While the time complexity of the two methods in one epoch is comparable ($O(Pn)$ and $O(n^2)+ O(Pn)$), the spectral method reduces the memory complexity from $O(PnB)$ to $O(n^2)+O(PB)$. Thus, when $P = O(n)$, both methods have a time complexity  $O(n^2)$, but \emph{the spectral method reduces the memory complexity by $B$}. When $P\gg n$, both methods again have the same time complexity $O(Pn)$ but \emph{the spectral method reduces the memory complexity by $n$}. Thus, with improved memory complexity, the proposed method scales effectively to both high-dimensional feature spaces and large sample sizes. In experiments, we observe clear advantages of the spectral method on high-dimensional datasets (DBCD, DLBCL, VDV, and LUNG1) in terms of both memory usage and computational efficiency. Additionally, experiments on counting-processes datasets with increasing numbers of samples confirm that our method scales with only mild increases in runtime and memory consumption (see Fig.~\ref{fig:memory_runtime} in Section~\ref{sec: experiments}).

\begin{table}[!t]
\begin{center}
\resizebox{\linewidth}{!}{
\begin{tabular}{ccccccc}
\hline
Dataset& Type & $n$ & $d$ &  Split  & Model &Censoring rate\\
\hline 
DBCD &   \multirow{4}{*}{SA}      &295 &4919&  90\%(5-fold CV)/10\% & MLP& 73.2\%\\
DLBCL    &    &240 &7399&  90\%(5-fold CV)/10\%  & MLP& 57.2\%\\
VDV     &      &78  &4705&  90\%(5-fold CV)/10\% & MLP& 43.5\%\\
 LUNG1 &  &422 &17M& 65\%/15\%/20\% &  CNN& 11.4 \%\\
\hline
ADS100 &\multirow{4}{*}{CP}   &100 &50 & 70\%/15\%/15\% &  MLP& 48.3\%\\
ADS1K &  &1K &50 & 70\%/15\%/15\% &  MLP& 48.3\%\\
ADS10K &  &10K &50 & 70\%/15\%/15\% &  MLP& 48.3\%\\
MovieLens &   &100K &200 & 70\%/15\%/15\% &  MLP& 25\%\\
\hline
\end{tabular}
}
\end{center}
\caption{ Overview of datasets and corresponding network structures explored.  The dataset type is either standard Survival Analysis (SA) or Counting Process (CP) arrival rate regression. $n$ denotes the number of samples and $d$ denotes data dimension. LUNG1 are CT scans  of dimension $68\times512\times 512$, leading to $d\simeq$17M features per sample. 
For smaller datasets,  we use 10\% as a hold-out test set and conduct $5$-fold cross-validation on the training set to determine hyperparameters. For larger datasets, we use fixed train/validation/test splits.}
\label{tab: Datasets}
\end{table}

\begin{table*}[!t]
\begin{center}
\begin{scriptsize}
\setlength{\tabcolsep}{3pt} 
\resizebox{0.9\textwidth}{!}{%
\begin{tabular}{cccccccccc}
\toprule
 &  & \textbf{DBCD} & \textbf{DLBCL} & \textbf{VDV} & \textbf{LUNG1} & \textbf{ADS100} & \textbf{ADS1k} & \textbf{ADS10k} & \textbf{MovieLens} \\
\textbf{Metrics} & \textbf{Algorithms} & \textbf{($d=4919$)} & \textbf{($d=7399$)} & \textbf{($d=4705$)} & \textbf{($d=17M$)} & \textbf{($d=50$)} & \textbf{($d=50$)} & \textbf{($d=50$)} & \textbf{($d=200$)}\\
\midrule
\multirow{7}{*}{\textbf{CI} $\uparrow$} 
& \textsc{DeepHit} \cite{lee2018deephit}     & 0.691±0.023 & 0.617±0.061 & 0.647±0.105 & 0.578 & -- & -- & -- & -- \\
& \textsc{DeepSurv}  \cite{Katzman2018DeepSurvPT} & 0.642±0.042 & 0.595±0.032 & 0.505±0.033 & 0.580 & 0.72 & 0.73 & 0.70 & 0.58\\
& \textsc{FastCPH}  \cite{yang2022fastcph}    & \textbf{0.724±0.109} & 0.574±0.069 & 0.567±0.065 & ---  & -- & -- & -- & -- \\
& \textsc{CoxTime} \cite{CoxTimeCC}          & 0.658±0.025 & 0.556±0.058 & 0.593±0.020 & \ding{56} & -- & -- & -- & -- \\
& CoxCC \cite{CoxTimeCC}            & 0.628±0.029 & 0.596±0.074 & 0.596±0.125 & \ding{56} & -- & -- & -- & -- \\
& \textsc{NMTLR} \cite{NMTLR}                & 0.716±0.023 & 0.577±0.045 & 0.536±0.139 & 0.539 & -- & -- & -- & -- \\
& \textsc{Spectral} (ours)                          & \textbf{0.724±0.007} & \textbf{0.625±0.065} & \textbf{0.716±0.050} & \textbf{0.633} & \textbf{0.75} & \textbf{0.77} & \textbf{0.73} & \textbf{0.603}\\
\midrule
\multirow{7}{*}{\textbf{AUC} $\uparrow$}
& \textsc{DeepHit} \cite{lee2018deephit}     & 0.728±0.031 & 0.617±0.061 & 0.647±0.105 & 0.653 & -- & -- & -- & -- \\
& \textsc{DeepSurv}  \cite{Katzman2018DeepSurvPT} & 0.661±0.044 & 0.667±0.037 & 0.505±0.033 & 0.622 & \textbf{0.65} & \textbf{0.66} & \textbf{0.66} & 0.533 \\
& \textsc{FastCPH}  \cite{yang2022fastcph}    & 0.677±0.131 & 0.609±0.097 & 0.576±0.070 & ---  & -- & -- & -- & -- \\
& \textsc{CoxTime} \cite{CoxTimeCC}          & 0.706±0.028 & 0.556±0.058 & 0.569±0.059 & \ding{56} & -- & -- & -- & -- \\
& CoxCC \cite{CoxTimeCC}            & 0.628±0.029 & 0.629±0.087 & 0.612±0.152 & \ding{56} & -- & -- & -- & -- \\
& \textsc{NMTLR} \cite{NMTLR}                & 0.608±0.033 & 0.585±0.041 & 0.423±0.214 & 0.55  & -- & -- & -- & -- \\
& \textsc{Spectral} (ours)                         & \textbf{0.734±0.016} & \textbf{0.698±0.076} & \textbf{0.762±0.046} & \textbf{0.697} & 0.63 & \textbf{0.66} & \textbf{0.66} & \textbf{0.535}\\
\midrule
\multirow{7}{*}{\textbf{RMSE} $\downarrow$}
& \textsc{DeepHit} \cite{lee2018deephit}     & 0.302±0.008 & 0.083±0.014 & 0.220±0.035 & 0.252 & -- & -- & -- & -- \\
& \textsc{DeepSurv}  \cite{Katzman2018DeepSurvPT} & 0.070±0.026 & 0.201±0.018 & 0.196±0.114 & 0.039 & 0.432 & 0.427 & 0.349 & 0.023 \\
& \textsc{FastCPH}  \cite{yang2022fastcph}    & 0.065±0.018 & 0.070±0.032 & 0.285±0.023 & --   & -- & -- & -- & -- \\
& \textsc{CoxTime} \cite{CoxTimeCC}          & 0.112±0.026 & 0.350±0.076 & 0.237±0.037 & \ding{56} & -- & -- & -- & -- \\
& CoxCC \cite{CoxTimeCC}            & 0.137±0.023 & 0.096±0.034 & 0.164±0.047 & \ding{56} & -- & -- & -- & -- \\
& \textsc{NMTLR} \cite{NMTLR}                & 0.124±0.031 & 0.208±0.062 & 0.206±0.053 & 0.104 & -- & -- & -- & -- \\
& \textsc{Spectral} (ours)                         & \textbf{0.052±0.018} & \textbf{0.065±0.045} & \textbf{0.093±0.060} & \textbf{0.037} & \textbf{0.41} & \textbf{0.405} & \textbf{0.269} & \textbf{0.022} \\
\bottomrule
\end{tabular}
}
\caption{Comparison of \textsc{Spectral} with state-of-the-art hazard models on 8 datasets: DBCD, DLBCL, VDV, LUNG1, ADS100, ADS1k, ADS10k, and MovieLens. Performance is evaluated in terms of CI (concordance index, $\uparrow$), AUC ($\uparrow$), and RMSE ($\downarrow$).  \ding{56} denotes an out-of-memory error and $-$ denotes that the method is not applicable. Standard deviations for the first three datasets are computed over 5-fold cross-validations. For LUNG1, ADS, and MovieLens which have fixed train-val-test splits, we provide the mean of $3$ runs.}
\label{tab:remaining_performance}
\end{scriptsize}
\end{center}
\end{table*}

\subsection{Extensions}
\label{sec: extensions}
The spectral method we proposed here for standard CoxPH is very versatile, and can be applied to several different survival analysis models:
\begin{itemize}
\item {\bf Weighted CoxPH \cite{Reweight2015, subgroupreweight, Buchanan2014Worthreweight}:} A common extension of CoxPH that weighs the hazard rate of each sample.
\item {\bf Heterogeneous Cox model \citep{hu2021subgroup}:} In this model, the set of samples partitioned into disjoint groups, and the hazard rate is regressed from class features in addition to per-sample features.
\item {\bf Deep Heterogeneous Hazard model (DHH):} Samples are again partitioned into classes that lack features; each class is given a different baseline hazard function $\lambda_0(\cdot)$.
\item \textbf{Accelerated Failure Time Model (AFT) \cite{Wei1992TheAF}:} In this model, the baseline hazard function $\lambda_0(\cdot)$ incurs a sample specific time-scaling distortion, also regressed from sample features.
\end{itemize}
For all of the above models, we can leverage Theorem~\ref{thm: ss} to apply our \textsc{Spectral} approach to accelerate the regression of model parameters  via MLE (see  App.~\arxiv{\ref{app: extensions}}{D~\cite{arxiv}}). Finally, our approach also readily generalizes to regresing counting process arrival rates in a manner akin to \citet{chen2023gateway}; we also describe this in App.~\arxiv{\ref{app: extensions}}{D~\cite{arxiv}}.
\section{Experiments}
\label{sec: experiments}

\subsection{Experiment Setup}
We summarize  our experimental setup; additional details  on datasets, algorithms and hyperparameters, metrics, and network structures are in App.~\arxiv{\ref{app: experiment details}}{G~\cite{arxiv}}. We make our code is publicly available.\footnote{\href{https://github.com/neu-spiral/SpectralSurvival}{\texttt{https://github.com/neu-spiral/SpectralSurvival}}}

\noindent{\bf Datasets.} We conduct experiments on 5 publicly available real-world datasets and 3 synthetic ones, summarized in Table~\ref{tab: Datasets} .  MovieLens and the synthetic ADS datasets (simulating advertisement clicking processes) are counting process arrival rate datasets, following the scenario of \citet{chen2023gateway}. Additional details about  each dataset are provided in App.~\arxiv{\ref{app: experiment details}}{G~\cite{arxiv}}. For smaller datasets, we use 10\% of each dataset as a hold-out test set and conduct $k$-fold cross-validation on the training set to determine hyperparameters, with $k$ indicated in Table~\ref{tab: Datasets}. For larger datasets, we use a training/validation/test split, as indicated in Table~\ref{tab: Datasets}. 

LUNG1 is a high-dimensional CT scan dataset, with voxels of dimensions $68\times512\times 512$, leading to $d\simeq$17M features per sample. It has been extensively studied by prior survival analysis works \cite{Haar2019, Brag2023, ZHENG2023109483, zhu2017wsisa}; all past works use significant dimensionality reduction techniques (see also Table~\ref{tab: LUNG1 legacy.}) to regress survival times. To make our analysis compareable, we use train/validation/test splits for LUNG1 following prior art \cite{zhu2017wsisa}.  To the best of our knowledge, our method \textsc{Spectral} is the first to successfully regress survival times in LUNG1 using the entire CT scans as inputs.

{\bf \noindent Survival Analysis Regression Methods.} We implement our spectral method (\textsc{Spectral}) and compare it against six SOTA survival analysis regression competitors. Four are continuous-time methods (\textsc{DeepSurv} \cite{Katzman2018DeepSurvPT}, \textsc{CoxTime} \cite{CoxTimeCC}, \textsc{CoxCC} \cite{CoxTimeCC}, \textsc{FastCPH} \cite{yang2022fastcph}), and regress the hazard rate from features.  Two are discrete-time methods (\textsc{DeepHit} \cite{lee2018deephit} and \textsc{NMTLR} \cite{NMTLR}); they split time into intervals and treat the occurrence of an event in an interval as a classification problem, thereby estimating the PMF. For all methods, except \textsc{FastCPH}, we use the DNN architectures shown in the last column Table.~\ref{tab: Datasets}: this is a 6-layer 3D CNN for LUNG1 data, and MLP with 2-6 layers, for the remaining datasets (see also App.~\arxiv{\ref{app: experiment details}}{G~\cite{arxiv}}); we treat the number of layers as a hyperparameter to be tuned. 
FastCPH has a fixed network structure, LassoNet~\cite{lassonet21}: we set the number of layers to 2-6 and again treat depth as a hyperparameter. For all the methods, we explore the same hyperparameter spaces. Additional implementation details are provided in App.~\arxiv{\ref{app: experiment details}}{G~\cite{arxiv}}.

Not all methods can be applied to the LUNG1, MovieLens, and ADS datasets.  LassoNet is incompatible with the 3D dataset LUNG1,  so we do not report experiments with \textsc{FastCPH}  on this dataset. As for MovieLens and ADS, they are counting processes-based datasets, where one sample may experience multiple events. Although it can be expressed as a nested partial likelihood, converting other losses to this setting is non-trivial. Thus, we only implement the  \textsc{DeepSurv}  baseline for these counting processes datasets.

\noindent\textbf{Legacy Methods.}  We also report the performance of four legacy methods executed on LUNG1 with extra techniques such as expert annotation (see Table~\ref{tab: LUNG1 performance.}); we do not re-execute these methods. We note that (a) three of these methods use additional, external features beyond the CT scans themselves and, again, (b) none can operate on the entire scan, but rely on dimensionality reduction methods to process their input. 

{\bf \noindent Hyperparameter Search.} Optimal hyperparameters are selected based on $k$-fold cross-validation or the validation set, as appropriate, using early stopping; we use CI as a performance metric (see below). The full set of hyperparameters explored is described in App.~\arxiv{\ref{app: experiment details}}{G~\cite{arxiv}}. 

{\bf \noindent Metrics.}
We adopted the 3 commonly used metrics for predictive performance: C-Index (CI, higher is better $\uparrow$), integrated AUC (AUC,  higher is better $\uparrow$), and RMSE (lower is better $\downarrow$). We also measure the runtime and memory consumption of each method. All five metrics are described in detail in App.~\arxiv{\ref{app: experiment details}}{G~\cite{arxiv}}.

\subsection{Experimental Results}
\label{subsec: results}
\noindent\textbf{Predictive Performance.}
Table.~\ref{tab:remaining_performance} shows  the predictive performance of all on eight datasets w.r.t.~CI, AUC, and RMSE. For high dimensional dataset LUNG1, \textsc{CoxCC} and \textsc{CoxTime} failed with  an out-of-memory error.  We observe that \textsc{Spectral} outperforms the baseline methods across all metrics and datasets except on ADS100, w.r.t. AUC. While higher CI and AUC indicate that \textsc{Spectral} has superior ranking performance (correctly predicting the order of events), higher RMSE demonstrates it also makes more accurate survival time predictions. The results suggest that leveraging \textsc{Spectral} to capture intrinsic hazard scores as the closed-form solution of the original partial likelihood improves predictive performance over existing survival analysis baselines.

\begin{table}[t]
\begin{center}
\resizebox{\linewidth}{!}{
\begin{tabular}{cccccc}
\toprule
Model&Algorithms &Features &  CI $\uparrow$& RMSE $\downarrow$  & AUC $\uparrow$\\
\midrule
2D-CNN & \textsc{DeepSurv} \cite{Haar2019}& Rad+Patches & 0.623*& $-$&0.64* \\
2D-CNN &  CE Loss\cite{Brag2023} &  Rad+Cli+Slices&  $-$ & $-$ &0.67* \\
3D-CNN &  Focal loss \cite{ZHENG2023109483} & Cli+ Cubes &$-$ & $-$ & 0.64*\\
2D-CNN & \textsc{DeepSurv} \cite{zhu2016deepconvsurv} & Patches &0.629*& $-$ & $-$\\
3D-CNN & {\sc Spectral} (ours) &CT (FB) &\textbf{0.633}&\textbf{0.037}  &\textbf{0.697}\\
\bottomrule
\end{tabular}}
\caption{Comparison to legacy methods applied to the Lung dataset from prior work \cite{Haar2019,Brag2023,ZHENG2023109483,zhu2016deepconvsurv}. $^*$ denotes values as reported by respective papers (no new execution).
 and $-$ denotes values not reported. Legacy methods use sub-sampling/dimensionality reduction approaches, operating on patches, cubes, or slices of the LUNG1 CT scans, but also have access to additional external features, such as extracted 2D radiomic (Rad) features and clinical (Cli) features (see also App.~\arxiv{\ref{app: experiment details}}{G~\cite{arxiv}}). Despite the latter advantages, by accessing the entire CT scan, \textsc{Spectral} outperforms prior art w.r.t.~predictive performance metrics reported. We note that \cite{Haar2019,Brag2023,ZHENG2023109483,zhu2016deepconvsurv} do not report memory or runtime performance.} 
\label{tab: LUNG1 legacy.}
\end{center}
\end{table}

\begin{figure*}[!t]
  \centering
  \subfloat[Memory of Type SA\label{subfig: mem sa}]{
    \includegraphics[width=0.25\textwidth]{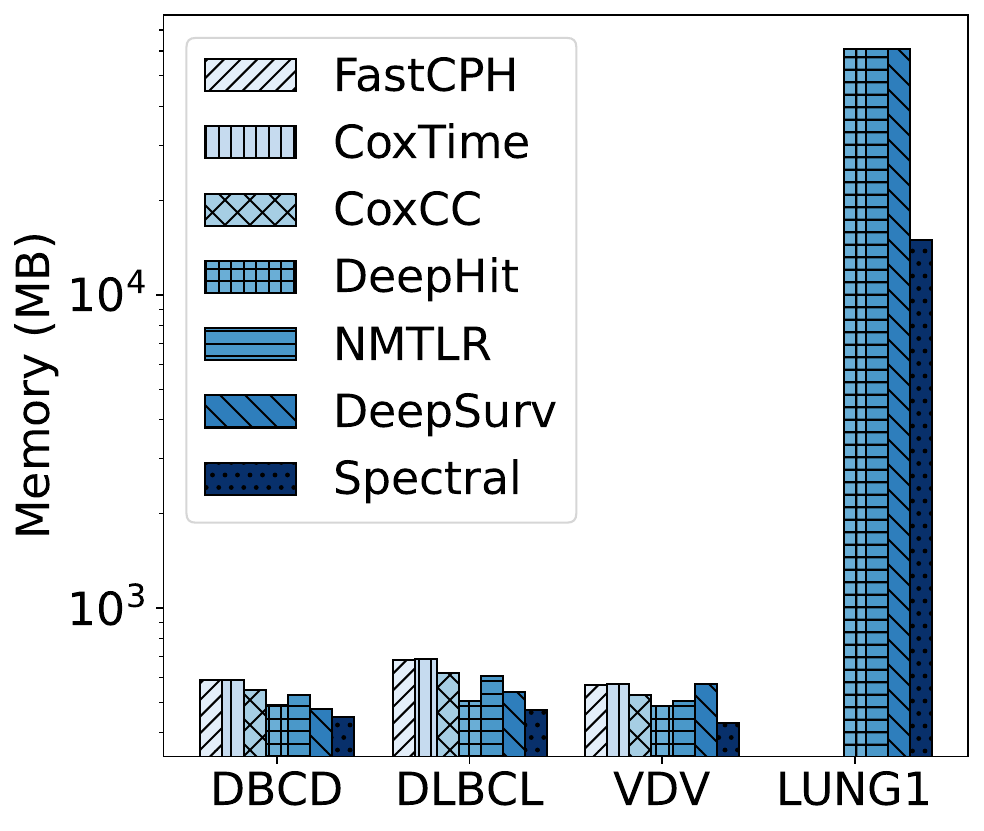}%
  }
  \hfill
  \subfloat[Memory of Type CT\label{subfig: mem ct}]{
    \includegraphics[width=0.25\textwidth]{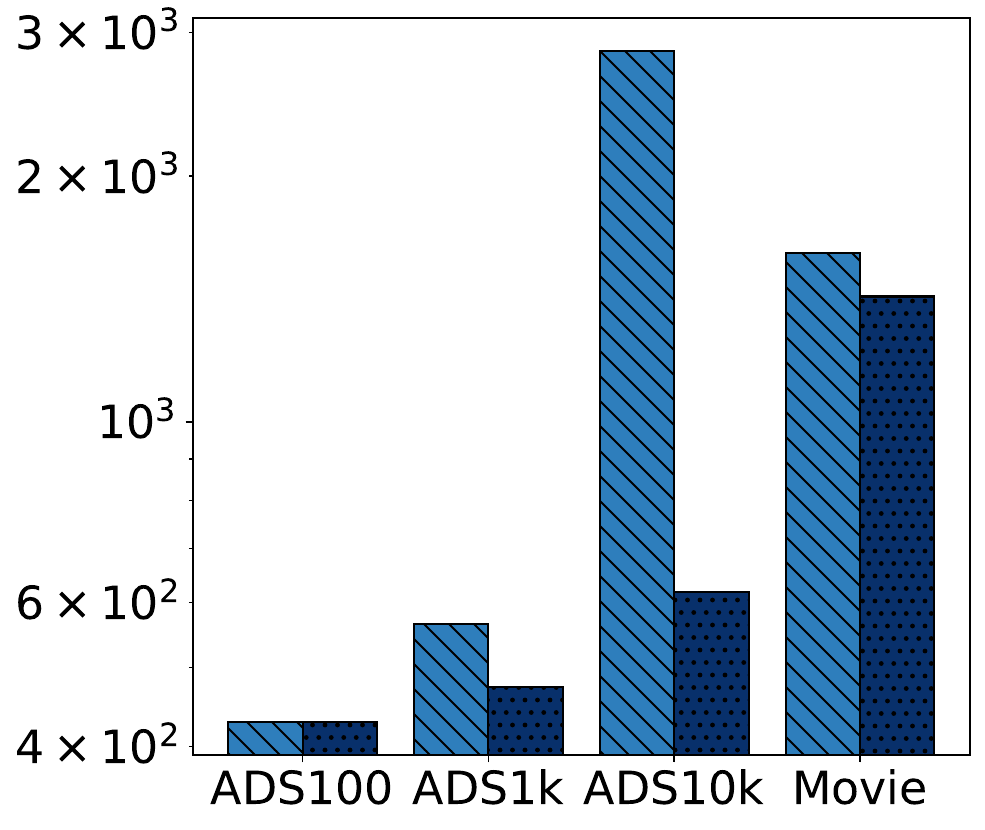}%
  }
  \hfill
  \subfloat[Runtime of Type SA\label{subfig: rt sa}]{%
    \includegraphics[width=0.25\textwidth]{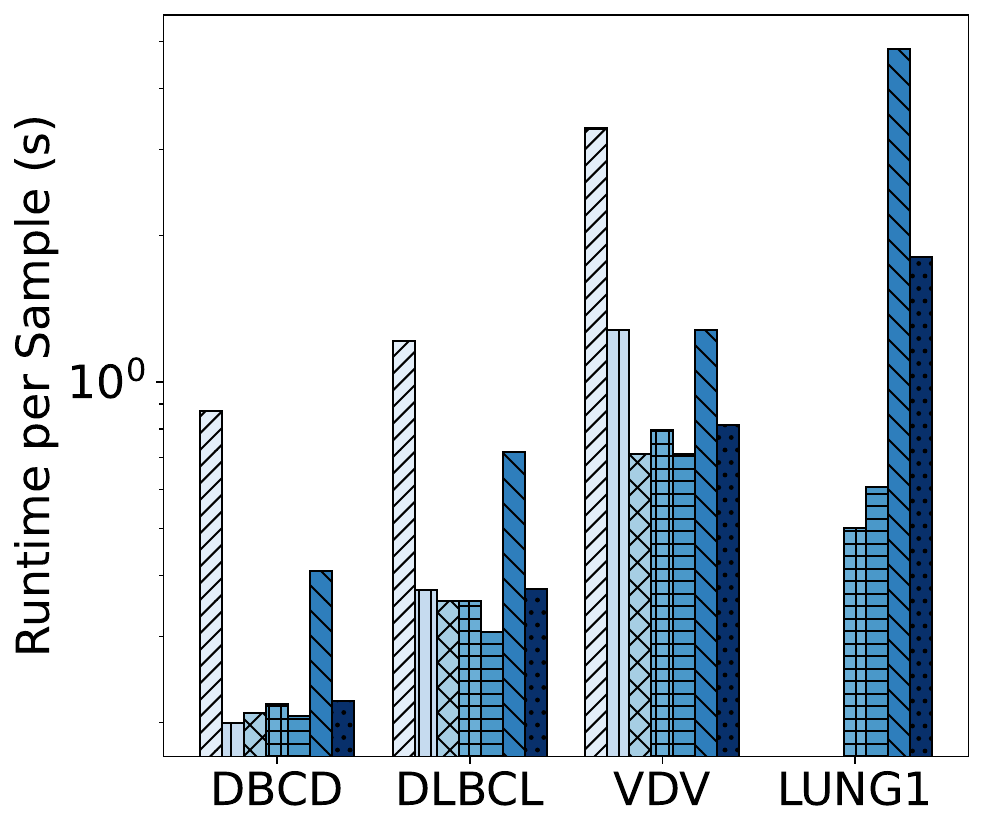}%
  }
  \hfill
  \subfloat[Runtime of TypeCT\label{subfig: rt ct}]{
    \includegraphics[width=0.25\textwidth]{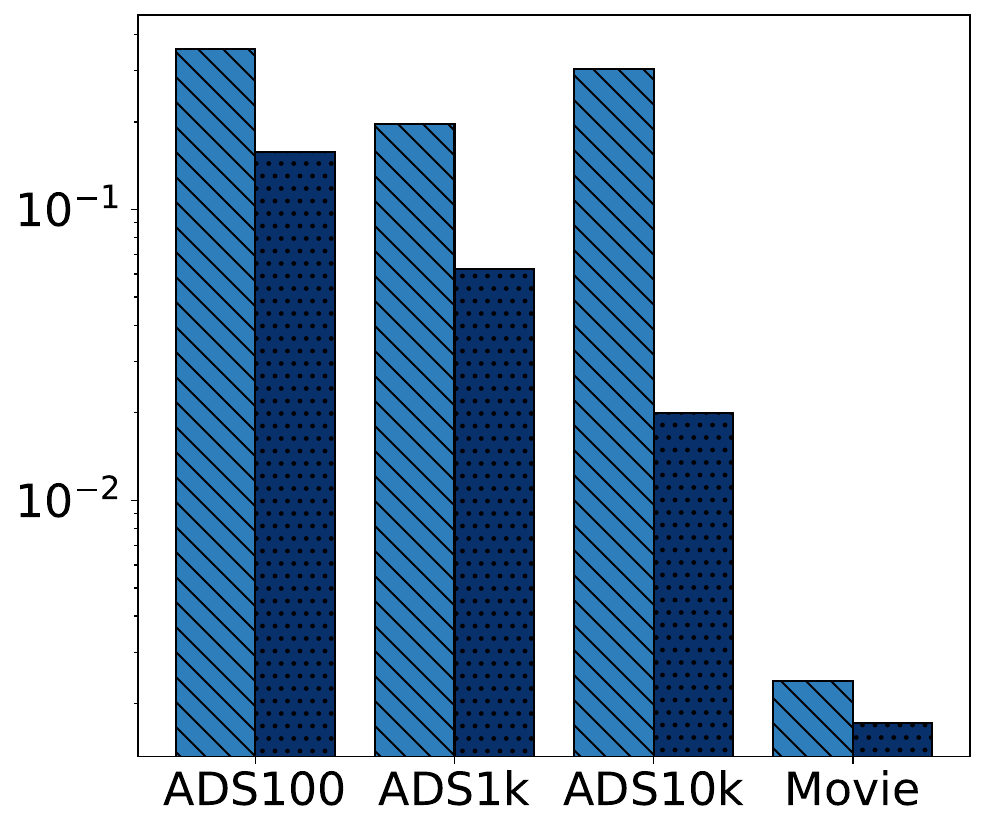}%
  }
  \caption{Comparison of memory (MB) and runtime per sample (s) between \textsc{Spectral} and competitor methods across survival analysis (SA)  and counting process (CP) datasets. All values are also reported in Table~\arxiv{\ref{tab: runtime and memory}}{10} in App.~\arxiv{\ref{app: additional experiments}}{H~\cite{arxiv}}. W.r.t.~memory, \textsc{Spectral} outperforms all other baselines in all cases. Moreover, the advantage is more obvious in the computational intensive cases: LUNG1 (high $d$) and ADS100K (high $n$). In terms of runtime per sample, \textsc{Spectral} consistently accelerates the convergence over its \textsc{DeepSurv}  base model, with which it shares the same objective. It has a comparable performance to other methods that use simpler objectives; however, these simpler methods suffer from worse predictive performance compared to \textsc{Spectral} and, often, \textsc{DeepSurv} (see also Table~\ref{tab:remaining_performance} and Fig.~\ref{fig:teaser_standard}).}
  \label{fig:memory_runtime}
\end{figure*}

\sloppy
Table~\ref{tab: LUNG1 legacy.} further contrasts the predictive performance of \textsc{Spectral} to legacy methods on the ultra high-dimensional dataset LUNG1, as reported in prior art. Legacy methods use sub-sampling/dimensionality reduction approaches like sampling random patches, while having access to additional clinical or radiomic information  \cite{zheng2023cancergroups, Brag2023, Haar2019} or utilizing patches extracted via pathologists' manual annotation~\cite{zhu2016deepconvsurv} (see also App.~\arxiv{\ref{app: experiment details}}{G~\cite{arxiv}}). By accessing the entire CT scan, \textsc{Spectral} outperforms all methods in predictive performance metrics reported.

\fussy


In Table~\ref{tab: LUNG1 performance.}, we further explore the impact of mini-batch execution on competitors. No competitor can be executed with full-batch due to the Siamese structure introduced by their losses. We thus ran each competitor with the largest mini-batch size possible (20\%)  that does not yield an out-of-memory error.  For comparison, though \textsc{Spectral} can be executed in full-batch, we also execute it with in  mini-batch mode in fitting the model to intrinsic scores. We observe that none of the baseline methods catch up to \textsc{Spectral} or event he legacy methods in Table~\ref{tab: LUNG1 legacy.} with mini-batch SGD. We attribute the performance gap to the biased gradient estimation (see App.~\arxiv{\ref{app: biased est}}{C~\cite{arxiv}}).



\begin{table}[t]
\begin{center}
\resizebox{\linewidth}{!}{
\begin{tabular}{cccccccc}
\hline
Model&Algorithms &Features &  CI $\uparrow$& RMSE $\downarrow$  & AUC $\uparrow$ & Memory& Time(s)\\
\hline
\multirow{8}{*}{3D-CNN}& \multirow{2}{*}{\sc DeepSurv} & CT (FB)  & \ding{56}&\ding{56}   & \ding{56} & $>$80G&\ding{56}\\
& & CT (20\%B) &  0.580& 0.039& 0.622& 61GB&2038\\
&\multirow{2}{*}{\sc DeepHit} & CT (FB)  & \ding{56} & \ding{56}  & \ding{56} & $>$80GB&\ding{56}\\
& & CT (20\%B) &  0.578& 0.252& 0.653& 61GB& 211\\
&\multirow{2}{*}{\sc MTLR} & CT (FB)  & \ding{56} & \ding{56} & \ding{56} & $>$80GB&\ding{56}\\
&  & CT (20\%B) & 0.539&0.104 & 0.55& 61GB& 256\\
&\multirow{2}{*}{\sc CoxCC} & CT (FB) & \ding{56} & \ding{56}  & \ding{56} & $>$80GB&\ding{56}\\
& & CT (20\%B) & \ding{56}&  \ding{56} & \ding{56} & $>$80GB& \ding{56}\\
&\multirow{2}{*}{\sc CoxTime} & CT (FB) & \ding{56} & \ding{56}   & \ding{56} & $>$80GB& \ding{56}\\
& & CT (20\%B) & \ding{56}  & \ding{56}  & \ding{56} & $>$80GB&\ding{56}\\
&{\sc Spectral} (ours) &CT (FB) &\textbf{0.633}&\textbf{0.037}  &\textbf{0.697}& \textbf{15GB}& 760\\
\hline
\end{tabular}}
\caption{Additional mini-batch experiments on LUNG1 dataset. We  compare \textsc{Spectral} with five SOTA hazard models applied to the entire CT scan inputs (bottom), using mini-batch SGD.  \ding{56} indicates out-of-memory failure. FB denotes full batch.  \textsc{Spectral} operates on the whole input, surpassing SoTA with multi-modal inputs including extracted patches with manual annotation.  All competitor methods run out of memory beyond $10\%$ mini-batches on an 80GB A100 GPU, while \textsc{Spectral} works under the full-batch regime, improving performance.  } 
\label{tab: LUNG1 performance.}
\end{center}
\end{table}

\noindent\textbf{Scalability Performance Comparison.}
To verify the scalability of the proposed \textsc{Spectral} w.r.t.~memory and runtime, we compare it with the baselines in two groups: (a) the Survival Analysis (SA) type datasets DBCD, DLBCL, VDV, and LUNG1 and (b) the Counting Process (CP)   ADSx and MovieLens.

We report memory performance for the two types in Figs.~\ref{subfig: mem sa} and~\ref{subfig: mem ct}, respectively. We observe that \textsc{Spectral} outperforms all other baselines in all cases. Moreover, the advantage is more obvious in the computational intensive cases: LUNG1 (high $d$) and ADS100K (high $n$). Note that, for LUNG1, none of the baselines can operate in full batch mode. Thus, we report the memory of the baseline methods with 20\% batch size. Nevertheless, all baseline methods consume $4\times$ the memory of \textsc{Spectral}. For ADS, textsc{Spectral} shows superior scalability over $n$ compared to \textsc{DeepSurv} . Particularly, although they require similar amount of memory for ADS100, \textsc{DeepSurv}  requires $5\times$ the memory compared to \textsc{Spectral} for ADS100K. Overall, this improved performance in terms of memory is instrumental in scaling \textsc{Spectral} over large (high $n$) and/or high-dimensional (large $d$), in contrast to competitors.


We report performance w.r.t.~running time per sample (i.e., running time divided by $n$) in Figs.~\ref{subfig: rt sa} and~\ref{subfig: rt ct}, respectively; we also  report absolute running time values in Table~\arxiv{\ref{tab: runtime and memory}}{10} in App.~\arxiv{\ref{app: additional experiments}}{H~\cite{arxiv}}.
We observe that \textsc{Spectral} consistently accelerates the convergence over its \textsc{DeepSurv}  base model, with which it shares the same objective (namely, the CoxPH loss). It has a comparable performance to other methods that use simpler objectives; we note that, in conjunction with the lack of scalability in terms of memory, these simpler methods suffer from worse predictive performance compared to \textsc{Spectral} and, often, \textsc{DeepSurv} (see also Table~\ref{tab:remaining_performance} and Fig.~\ref{fig:teaser_standard}).

\noindent\textbf{Sensitivity to Hyperparameters.}
We evaluate the sensitivity of the proposed method to the ADMM parameter $\rho$ and the maximum iterations allowed for the power method. As shown in Table~\ref{tab:rho-sensitivity}, the model performs best around the standard setting $\rho=1$, as suggested in prior work~\cite{yildiz21a}, achieving the highest AUC and CI with the lowest RMSE. We tested a range of values ${0.1, 0.5, 1, 2, 5, 10}$ and found that performance drops noticeably for both smaller and larger $\rho$. Moreover, we observe that setting $\rho=1$ leads to optimal performance for most datasets. In terms of inner-loop optimization, for stable and competitive results, we set the maximum number of power iterations to at least 50. Consequently, we observe that the power method converges efficiently across all datasets—typically within 2 to 50 rounds. The first round requires 50 iterations, while subsequent rounds usually converge in 2 iterations. 

\begin{table}[!t]
\centering
\begin{tabular}{cccc}
\toprule
$\boldsymbol{\rho}$ & \textbf{AUC} & \textbf{CI} & \textbf{RMSE} \\
\midrule
0.1 & 0.70 $\pm$ 0.05 & 0.64 $\pm$ 0.05 & 0.22 $\pm$ 0.04 \\
0.5 & 0.69 $\pm$ 0.12 & 0.69 $\pm$ 0.09 & 0.08 $\pm$ 0.04 \\
1 & \textbf{0.77 $\pm$ 0.04} & \textbf{0.70 $\pm$ 0.03} & \textbf{0.11 $\pm$ 0.01} \\
2 & 0.63 $\pm$ 0.09 & 0.70 $\pm$ 0.04 & 0.17 $\pm$ 0.02 \\
5 & 0.56 $\pm$ 0.05 & 0.56 $\pm$ 0.04 & 0.11 $\pm$ 0.05 \\
10 & 0.43 $\pm$ 0.10 & 0.37 $\pm$ 0.08 & 0.17 $\pm$ 0.02 \\
\bottomrule
\end{tabular}
\caption{Sensitivity of the \textsc{Spectral} to the ADMM penalty parameter $\rho$ on DBCD.}
\label{tab:rho-sensitivity}
\end{table}

\noindent\textbf{Depth and Batch Size.}
\textsc{Spectral} also scales more gracefully with increasing depth, exhibiting only mild growth in runtime and memory usage. As shown in Table~\ref{tab:depth-scalability}, this efficiency stems from avoiding Siamese network structures, which impose higher computational overhead in baselines like \textsc{DeepSurv}.

Lastly, we evaluate how model performance is affected by mini-batch training. As shown in Table~\ref{tab:batch-size-combined}, standard survival analysis methods relying on mini-batches to scale (such as DeepHit and DeepSurv) exhibit a consistent drop in both CI and AUC when the batch size is reduced—particularly below $10\%$. In contrast, our spectral method remains stable across all batch sizes, as we solve the full partial likelihood optimization problem and convert the survival analysis problem into regression problem. This again highlights the advantage of solving the full partial likelihood directly.
\begin{table}[!t]
\centering
\resizebox{\linewidth}{!}{
\begin{tabular}{lcccccc}
\toprule
\textbf{Depth} & \textbf{Algorithm} & \textbf{AUC} & \textbf{CI} & \textbf{RMSE} & \textbf{Time (s)} & \textbf{GPU Mem (MiB)} \\
\midrule
2  & Spectral   & 0.77 & 0.77 & 0.11 &  67 & 451 \\
10 & Spectral   & 0.74 & 0.66 & 0.04 &  85 & 459 \\
50 & Spectral   & 0.52 & 0.53 & 0.09 & 116 & 499 \\
\midrule
2  & DeepSurv   & 0.64 & 0.64 & 0.10 & 116 & 469 \\
10 & DeepSurv   & 0.62 & 0.58 & 0.04 & 128 & 485 \\
50 & DeepSurv   & 0.56 & 0.56 & 0.04 & 190 & 619 \\
\bottomrule
\end{tabular}
}
\caption{Scalability of Spectral and DeepSurv with respect to network depth on DBCD.}
\label{tab:depth-scalability}
\end{table}

\begin{table}[!t]
\centering
\resizebox{\linewidth}{!}{
\begin{tabular}{llcccccccc}
\toprule
\multirow{2}{*}{\textbf{Model}} & \multirow{2}{*}{\textbf{Metric}} 
& \multicolumn{4}{c}{\textbf{Lung1 (Batch Size \%)}} 
& \multicolumn{4}{c}{\textbf{DLBCL (Batch Size \%)}} \\
\cmidrule(lr){3-6} \cmidrule(lr){7-10}
& & 100 & 10 & 5 & 1 & 100 & 50 & 10 & 5 \\
\midrule
\multirow{2}{*}{Spectral} 
& CI  & 0.63 & 0.60 & 0.61 & 0.60 & 0.63 & 0.51 & 0.60 & 0.61 \\
& AUC & 0.70 & 0.63 & 0.65 & 0.65 & 0.70 & 0.69 & 0.66 & 0.63 \\
\midrule
\multirow{2}{*}{DeepHit}  
& CI  & --   & 0.58 & 0.51 & 0.53 & 0.62 & 0.60 & 0.47 & 0.52 \\
& AUC & --   & 0.65 & 0.54 & 0.49 & 0.62 & 0.65 & 0.54 & 0.55 \\
\midrule
\multirow{2}{*}{DeepSurv} 
& CI  & --   & 0.58 & 0.55 & 0.51 & 0.60 & 0.55 & 0.57 & 0.56 \\
& AUC & --   & 0.62 & 0.57 & 0.54 & 0.67 & 0.57 & 0.58 & 0.60 \\
\bottomrule
\end{tabular}
}
\caption{Performance (CI and AUC) of different methods under varying batch sizes on \textbf{Lung1} and \textbf{DLBCL}. ``--'' indicates training failure due to out-of-memory (OOM).}
\label{tab:batch-size-combined}
\end{table}

\section{Conclusion}
We have shown how a spectral method motivated by ranking regression can be applied to scaling survival analysis via {\sc CoxPH} and multiple variants. Extending this model to other variants of {\sc CoxPH} and \textsc{DeepSurv}  is an interesting open direction. Our \textsc{Spectral} method outperforms SOTA methods on various high-dimensional medical datasets such as genetic and cancer datasets and counting processes datasets such as MovieLens; given the importance of survival analysis, our method may more broadly benefit the healthcare field and commercial advertisements, opening the application of survival analysis methods to ultra-high dimensional medical datasets. Finally, although \textsc{Spectral} shows significant improvement over SOTA in complex datasets, this is not as marked in the low-dimensional regime; we leave exploring this as for future work. 

\begin{acks}
The authors gratefully acknowledge support from the National Science Foundation (grants {2112471} and {1750539}).
\end{acks}

\bibliographystyle{ACM-Reference-Format}
\bibliography{sample-base}

\arxiv{
\appendix
\onecolumn

\section{Non-Parametric and Parametric Survival Analysis Methods}
\label{app: related work}
Survival analysis is a well-established discipline: we refer the interested reader to Aelen et al \cite{AalenBook} for a more thorough exposition on the subject. For completeness but, also, to motivate our focus on the Cox Proportional Hazard model (itself a semi-parametric method, described in Sec.~\ref{sec: preliminaries}), we provide technical overview of  parametric and non-parametric survival analysis methods in this section.

\subsection{Notation}

We briefly restate the notational conventions from Sec.~\ref{sec: preliminaries}. We consider a dataset  $\mathcal{D}=\{\vx_i,\obt_i,\Delta_i\}^n_{i=1},$ comprising $n$ samples, each associated with a $d$-dimensional feature vector $\vx_i \in \R^d$. Each sample is additionally associated with an \emph{event time} $\et_i>0$ and a \emph{censoring time} $\ct_i>0$ (see Fig.~\ref{fig: illustration}); then, $O_i=\min(T_i,C_i)$ is the \emph{observation time} (either event or censoring) and $\Delta_i\equiv \ind_{\et_i\leq \ct_i}$ is the \emph{event indicator}, specifying whether the event was indeed observed.

 The goal of survival analysis is to regress the distribution of event times from event features. The distribution is typically either modeled via the survival function or the hazard function. The survival function is the complementary c.d.f. of event times, i.e.,
$$S(t|\vx)=P(T>t|\vx).$$  Given a sample $\vx$, the hazard function is:
\begin{align*}
    \lambda(t|\vx)=\lim_{\delta \to 0} \frac{P(t\leq\! T\!\leq t\!+\!\delta|T\geq t,\vx)}{\delta}=\frac{f(t|\vx)}{S(t|\vx)}=-\frac{S'(t|\vx)}{S(t|\vx)},
\end{align*}
where $f(t|\vx)=-S'(t|\vx)$ is the density function of $\et$ given $\vx$. We can express the survival function via the cumulative hazard via
 $   S(t|\vx)=\exp{\left(-\int_0^t \lambda(s|\vx)\, ds\right)}.$

\subsection{Non-Parametric Methods}\label{app:non-parametric}

We first  review two non-parametric survival analysis methods. Non-parametric methods estimate one-dimensional survival time density without making distributional assumptions. The estimators below are akin, in principle, to traditional density estimation techniques via, e.g., binning and univariate histogram estimation~\cite{scott2015multivariate}, while kernel smoothing variants also exist (see, e.g., Eq.~\eqref{eq: smoothing NA} below). 

Like other non-parametric methods, the Kaplan-Meier and  Nelson-Aalen estimators do not explicitly capture the effects of features, limiting their ability to generalize out-of-sample. Na\"ively transferring them to a multi-variate would suffer from a curse-of-dimensionality~\cite{scott2015multivariate}, which motivates the introduction of parametric and semi-parametric methods.

\noindent{\bf Kaplan-Meier Estimator.}
The Kaplan-Meier estimator \cite{Kaplan1992}, also known as the product-limit estimator, is a non-parametric model estimating the survival function. The estimator of the survival function $S(t)$ in Eq.~\eqref{eq: hazard} is given by
\begin{align}
\hat{S}(t) = \prod_{t_i \leq t} \left(1 - \frac{d_i}{r_i}\right),
\end{align}
where: $ t_i\in \mathbf{R_+}$ is the ordered event times,$$d_i = \sum_{j=1}^{n} \ind_{O_j = O_i \cap\Delta_j = 1}$$ is the number of events, and  $$r_i = \sum_{j=1}^{n} \ind_{O_j \geq O_i}$$ is number of subjects at risk at time $t_i $.

\noindent{\bf Nelson-Aalen Estimator.} The Nelson-Aalen estimator \cite{NelsonAalen1978} is a non-parametric estimator used to estimate the cumulative hazard function in survival analysis. Formally, forthe set of samples $[n]$, let 
$$\mathcal{R}(t)=\{j\in[n]: t<O_j\}$$ be the set of at-risk samples at time $t$, and denote by $$Y(t)=|\mathcal{R}(t)|$$ its cardinality. Nelson-Aalen estimates the cumulative hazard $\Lambda(t)=\int_0^t \lambda(s)\, ds$ with the at-risk samples as
$$\hat{\Lambda}(t)=\sum_{j:T_j\leq t}\frac{1}{Y(T_j)}.$$

\noindent{\bf Kernel Smoothing.}
Kernel-smoothing can be applied to the Nelson-Aalen estimator, using the fact that   the derivative of the cumulative hazard rate directly leads to hazard rate estimates ~\cite{wang2005smoothing,AalenBook}.
In detail, given a smoothing kernel $K(\cdot)$ s.t. $\int_{t\in\R} K(t)dt=1$, the hazard function $\lambda(t)$ can be estimated as:
\begin{align}\label{eq: smoothing NA}
    \hat{\lambda}(t)=\frac{1}{b}\sum_{j:T_j\in \left[t-b, t+b\right]} K\left(\frac{t-T_j}{b}\right) \frac{1}{Y(T_j)},
\end{align}
where $b$ is the so-called bandwidth of the smoothing. Intuitively, the kernel smoothing is equivalent to taking the derivative of the cumulative hazard rate.

\subsection{Parametric Methods}
 Parametric method typically assume that survival times conditioned on features $\vx_i\in\mathbb{R}^d$  follow  distribution density $f(\cdot; \vtheta,\vx_i)$, parameterized by $\vtheta\in \mathbb{R}^m$. Estimating $\theta$ can be done through maximum likelihood estimation.  
In particular, assuming we have a dataset $\mathcal{D}$ containing $n$ independent individuals, with the $i$-th individual’s survival time denoted by $T_i$, covariate vector $\vx_i$, and censoring indicator $\Delta_i$, the  (full) likelihood of observations in $\mathcal{D}$ is given by: 


$$ \mathtt{FL}(\theta,\mathcal{D}) = \prod_{i=1}^{n} \left[ f(T_i; \theta, x_i) \right]^{\Delta_i} \left[ S(T_i; \theta, x_i) \right]^{1-\Delta_i} ,$$

To model the distribution (and thus, the full likelihood) it is common \citep{AalenBook} to assume that the underlying baseline survival time satisfies certain distribution (e.g., exponential \cite{WITTEN19927, jenkins2005survival}, Weibull~\cite{carroll2003use}, log-normal~\cite{Wei1992TheAF}, etc.): the regression function then regresses the parameters of the distribution from features via a linear or a deep model. On one hand, reducing estimation to few parameters of a well-known distribution reduces variance and makes the model more interpretable. On the other hand, it may increase bias, as the true underlying distribution may deviate from the one postulated by the model.

An alternative is to model the density function $f(\cdot)$ itself as a neural network~\citep{rindt2022sumo}. For example, Sumo-Net \cite{rindt2022sumo} uses a monotonic neural network \citep{mono-v124-chilinski20a} consisting of tanh functions, which could be considered as a smoother variant of step functions used in binning; as such, similar curse of dimensionality issues as non-parametric methods arise in this setting.

\section{Spectral Methods}
\label{app: spectral}
\subsection{Spectral Method for the Plackett-Luce Model}
For completeness, we first discuss the \textsc{Spectral} method introduced by Maystre and Grossglauser \cite{maystre2015fast}.  Given $n$ items, a dataset consisting of $m$ query sets $A_\ell\subseteq [n]$ can be described as $\mathcal{D}=\{(c_\ell, A_\ell)\}_{\ell=1}^m$, where $c_\ell\in A_\ell$ is the maximal choice (i.e., the winner) within set $A_\ell$, selected by a so-called max-oracle. A tuple  $(c_\ell, A_\ell)$ is termed an \emph{observation}.

The Plackett-Luce model \cite{PlackettLuce1975} assumes that each sample $i$ is associated with a non-negative score $\pi_i>0$. As the observations are independent, we can express the probability of each max-choice query as
\begin{align*}
    P(c_\ell| A_\ell, \vpi) = \frac{\pi_{c_\ell}}{\sum_{j\in A_\ell} \pi_j}=\frac{\pi_\ell}{\sum_{j\in A_\ell} \pi_j},
\end{align*}
where $\vpi\in R_+^n$ is the collection of all the scores and by abusing the notation. Abusing notation, for brevity, they denote the chosen score as $\pi_{\ell}\equiv \pi_{c_\ell}$.

This model can be used for so-called \emph{ranking observations}. Suppose there are $k_\ell$ items within the $\ell$-th observation, and they are ordered as $\sigma(1) \succ \dots \succ \sigma(k
_\ell)$. The partial likelihood of the ranking can be modeled via the Placket-Luce model as
\begin{align}
    P(\sigma(1) \succ \dots \succ \sigma(k_i))=\prod_{r=1}^{k_\ell} \frac{\pi_{\sigma(r)}}{\sum_{p=r}^{k_\ell} \pi_{\sigma(p)}}.
\end{align}
Hence, the total ranking probability is equivalent to  $k_\ell-1$ independent maximal-oracle queries, in which the winner is the $i$-th item over the remaining items. Hence, learning from rankings can be reduced to learning from max-oracle queries.

The log-likelihood of the Plackett-Luce model of $\pi$ over observation $\mathcal{D}$ becomes
\begin{align}\label{eq: placket-luce}
    \log \loss(\vpi|\mathcal{D})=\sum_{\ell=1}^m \left(\log\sum_{j\in \mathcal{A}_\ell} \vpi_j- \log  \vpi_i  \right).
\end{align}

Thus, to maximize the log-likelihood, Maystre and Grossglauser take the gradient of \eqref{eq: placket-luce} and set it to 0, yielding equations:
\begin{align}
    \frac{\partial\log \loss(\vpi|\mathcal{D})}{\partial \pi_i}=0, \forall i\in [n].
\end{align}
By Theorem.~1 in \cite{maystre2015fast}, the above set of equations is equivalent to:
\begin{align}
    \sum_{j\neq i} \pi_j \lambda_{ji}(\vpi) = & \sum_{j\neq i} \pi_i \lambda_{ij}(\vpi), \forall i \in [n],\\
    \lambda_{ji}=&\sum_{\ell\in W_i\cap L_j} \frac{1}{\sum_{t\in A_\ell} \vpi_t},\nonumber\\
W_i=\{\ell|i\in & A_\ell, i=c_\ell\},L_i=\{\ell|i\in R_{\ell},i \neq c_\ell\}, \nonumber
\end{align}
which indicates the optimal scores $\vpi$ is the steady state distribution of a Markov Chain defined by the transition rate $\lambda$. Hence, the optimal solution can be obtained by  iteratively computing the steady state and adapting the transition rates (which depend on $\vpi$), as indicated in Eq.~\eqref{eq:ssd}.

\subsection{Spectral Ranking Regression}
 Yildiz et al.~\cite{yildiz21a} introduce spectral ranking regression, extending the spectral ranking method of Maystre and Grossglausser to a setting where items have features. The parameters of the Plackett-Luce model are then regressed from these features.  where they serve the partial likelihood as the objective while having the neural network fitting the scores $\vpi$ as constraints from features. 
Thus, they propose the constrained ranking problem
 \begin{align}\label{eq: yildiz problem}
     \min_{\vpi, \mtheta } \loss(\vpi,\mtheta) =  \sum_{i=1}^n  &\log\sum_{j\in \mathcal{R}_i} (\vpi_j- \log  \vpi_i) \\
     s.t. \quad \pi_i = & h_{\mtheta}(\vx_i), \vpi\geq 0,\nonumber
 \end{align}
where $h_\theta$ is a neural network parameterized by $\theta$. Henceforth, the constrained optimization problem is solved by augmented ADMM with KL divergence as
\begin{align}\label{eq: spectral lagrangian}
    \loss(\vpi,\mtheta,\vu)\equiv \sum_{i=1}^n \log\sum_{j\in \mathcal{R}_i} (\vpi_j- \log  \vpi_i) 
    +\vu^\top (\vpi - \tilde{h}_{\mtheta}(\mx))+\rho \mathcal{D}_{KL}(\vpi||h_{\mtheta}(\mx)).
\end{align}

By setting the derivative of the augmented ADMM loss to 0, 
\begin{align}
     \frac{\partial \loss(\vpi,\mtheta,\vu)}{\partial \vpi_i}
        =&\sum_{\ell\in W_i} \left(\frac{1}{\sum_{t\in \mathcal{R}_\ell} \vpi_t}- \frac{1}{\vpi_i}   \right)+\sum_{\ell\in L_i} \frac{1}{\sum_{t\in \mathcal{R}_\ell} \vpi_t} +\vu^{k}_i\\
        &+ \rho \frac{\partial D_{KL} (\vpi||\tilde{h}_{\mtheta}(\mx))}{\partial \vpi_i}=0\nonumber
\end{align}
Yildiz et al. proved that the stationary solution of the ADMM version of partial likelihood remains the steady state of a Markov chain. 
\begin{theorem}[Yildiz et al. \cite{yildiz21a} \label{thm: yildiz}] Given $\tilde{\pi_i}^k=h_{\mtheta}(\vx_i)\in \mathbb{R_+}^n$ and $\vy^k \in \mathbb{R}^n$, a stationary point $\vpi$ of the Augmented Lagrangian \eqref{eq: spectral lagrangian} satisfies the balancing equation of a continuous-time Markov Chain with transition rates:
    \begin{align}
    \psi_{ji}(\vpi)= \begin{cases}
\mu_{ji}+\frac{2\vpi_i \sigma_i(\vpi) \sigma_j(\vpi)}{\sum_{t\in [n]_-}\vpi_t \sigma_t(\vpi)-\sum_{t\in [n]_+}\vpi_t \sigma_t(\vpi)} & \textit{if } j \in [n]_+\land i\in [n]_- \\
\mu_{ji} & \textit{otherwise},\\
\end{cases}
\end{align}
with $[n]_{+}=\{i:\sigma_i(\vpi)\geq 0\}$, $[n]_{-}=\{i:\sigma_i(\vpi)\leq 0\}$, and

\begin{align*}
\sigma_i(\vpi)= &\rho \frac{\partial D_{KL}(\vpi||\tilde{h}_{\mtheta}(\mx))}{\partial \vpi_i}+u^{(k)}_i\\
\mu_{ji}=&\sum_{\ell\in W_i\cap L_j} \frac{1}{\sum_{t\in A_\ell}  \vpi_t}, \\
W_i=\{\ell|i\in & A_\ell, i=c_\ell\},L_i=\{\ell|i\in A_{\ell},i \neq c_\ell\}.
\end{align*}
\end{theorem}
Yildiz et al. use this theorem to combine ADMM with a \textsc{Spectral} method, as outlined in steps 
\eqref{eq:admm}.
Our Theorem~\ref{thm: ss} generalizes  Theorem~\ref{thm: yildiz} as (a) it considers the censoring and, most importantly,  (b) includes weights. The latter enables the application of the \textsc{Spectral} methods by Maystre and Grossglausser and Yildiz et al. to a broad range of survival analysis models.

\section{Biased Estimation with Mini-batches}
\label{app: biased est}
In this section, we prove that the mini-batched partial likelihood is a biased estimator of the original partial likelihood and we provide the condition of equivalence. 

For completeness, we recall the the partial likelihood is defined as:

\begin{align*}
    \mathcal{L}_{\mathtt{NLL}}(\dataset,\mtheta) = & \frac{1}{n} \sum_{i=1}^n\Delta_i \left(\log \sum_{j\in \mathcal{R}(T_i)} e^{\mtheta^\top \vx_j}-\mtheta^\top \vx_i\right)
\end{align*}

Therefore, the corresponding mini-batch estimation is as below

\resizebox{\columnwidth}{!}{
\begin{minipage}{\columnwidth}
\begin{align*}
    & \mathcal{L}_{\mathtt{NLL-mini batch}}(\dataset,\mtheta) \\
    = & \frac{1}{|\mathcal{B}|}\sum_{i\in \mathcal{B}} \Delta_i \left(\log \sum_{j\in \mathcal{R_{B}}(T_i)} e^{\mtheta^\top \vx_j}-\mtheta^\top \vx_i\right)\\
    = & \frac{1}{|\mathcal{B}|}\sum_{i=1}^n \mathbb{I}\left[i\in \mathcal{B}\right] \Delta_i \left(\log \sum_{j\in \mathcal{R}(T_i)} \mathbb{I}\left[j\in \mathcal{B}\right]  e^{\mtheta^\top \vx_j}-\mtheta^\top \vx_i\right),
\end{align*}
\end{minipage}
}

where $\mathcal{B} \subseteq [n]$ denotes the batch and the batch risk set is denoted as $\mathcal{R}_{\mathcal{B}}(t)\equiv \{j\in \mathcal{B}: t<O_j\}$.

Thus, take CoxPH model for example, we show how it is biased in Lemma.~\ref{lem: biased loss}.

\begin{lem}\label{lem: biased loss}
    Given dataset $\dataset$ and a CoxPH model parameterized by $\vtheta$, the bias introduced by the mini-batch $\mathcal{B}$ is
    \begin{displaymath}
        \mathbb{E}_\mathcal{B}\left[\mathcal{L}_{\mathtt{NLL}}(\dataset,\mtheta) - \mathcal{L}_{\mathtt{NLL-mini batch}}(\dataset,\mtheta) \right] = -  \frac{1}{n}\sum_{i=1}^n \Delta_i \mathbb{E}_{\mathcal{B}|i\in \mathcal{B}}\left[\log \left(1- \frac{\sum_{j\in \mathcal{R}(T_i) } \mathbb{I}\left[j\notin \mathcal{B}\right]e^{\mtheta^\top \vx_j}}{\sum_{j\in \mathcal{R}(T_i) } e^{\mtheta^\top \vx_j}}\right)\right],
    \end{displaymath}
where $\mathbb{I}\left[j\notin \mathcal{B}\right]$ is the indicator function of $j$ not being in the batch.
\end{lem}

\begin{proof}
First, we can express the difference between the original partial likelihood and its mini-batch variants as

\begin{align*}
        &\mathbb{E}_\mathcal{B}\left[\mathcal{L}_{\mathtt{NLL}}(\dataset,\mtheta) - \mathcal{L}_{\mathtt{NLL-mini batch}}(\dataset,\mtheta) \right]\\
    =& \frac{1}{n} \sum_{i=1}^n\Delta_i \left(\log \sum_{j\in \mathcal{R}(T_i)} e^{\mtheta^\top \vx_j}-\mtheta^\top \vx_i\right) - \mathbb{E}_\mathcal{B}\left[ \frac{1}{|\mathcal{B}|}\sum_{i=1}^n \mathbb{I}\left[i\in \mathcal{B}\right] \Delta_i \left(\log \sum_{j\in \mathcal{R_{B}}(T_i)} e^{\mtheta^\top \vx_j}-\mtheta^\top \vx_i\right)\right].
\end{align*}

As every sample is selected with probability $P(i\in \mathcal{B})=\frac{|\mathcal{B}|}{n},$
the second terms within both brackets cancels. Then, the equation above becomes
\begin{align*}
    & \frac{1}{n} \sum_{i=1}^n\Delta_i \log \sum_{j\in \mathcal{R}(T_i)} e^{\mtheta^\top \vx_j} -  \frac{1}{|\mathcal{B}|}\sum_{i=1}^n \mathbb{E}_\mathcal{B}\left[\mathbb{I}\left[i\in \mathcal{B}\right] \Delta_i \log \sum_{j\in \mathcal{R_{B}}(T_i)} \mathbb{I}\left[j\in \mathcal{B}\right] e^{\mtheta^\top \vx_j}\right].
\end{align*}
Then, by conditioning and manipulating the terms, we have 
\begin{align*}
    =& \frac{1}{n} \sum_{i=1}^n\Delta_i \log \sum_{j\in \mathcal{R}(T_i)} e^{\mtheta^\top \vx_j} -  \frac{1}{|\mathcal{B}|}\sum_{i=1}^n P(i\in \mathcal{B})\mathbb{E}_{\mathcal{B}|i\in \mathcal{B}}\left[\Delta_i \log \sum_{j\in \mathcal{R}(T_i)} \mathbb{I}\left[j\in \mathcal{B}\right]e^{\mtheta^\top \vx_j}\right]\\
     =& \frac{1}{n} \sum_{i=1}^n\Delta_i \log \sum_{j\in \mathcal{R}(T_i)} e^{\mtheta^\top \vx_j} -  \frac{1}{n}\sum_{i=1}^n \mathbb{E}_{\mathcal{B}|i\in \mathcal{B}}\left[\Delta_i \log \left( \sum_{j\in \mathcal{R}(T_i) } \mathbb{I}\left[j\in \mathcal{B}\right]e^{\mtheta^\top \vx_j}\right) \right]\\
    =& -  \frac{1}{n}\sum_{i=1}^n \mathbb{E}_{\mathcal{B}|i\in \mathcal{B}}\left[\Delta_i \log \frac{ \sum_{j\in \mathcal{R}(T_i)} \mathbb{I}\left[j\in \mathcal{B}\right]e^{\mtheta^\top \vx_j}}{\sum_{j\in \mathcal{R}(T_i)} e^{\mtheta^\top \vx_j}} \right]\\
    =& -  \frac{1}{n}\sum_{i=1}^n \Delta_i \mathbb{E}_{\mathcal{B}|i\in \mathcal{B}}\left[\log \left(1- \frac{\sum_{j\in \mathcal{R}(T_i) } \mathbb{I}\left[j\notin \mathcal{B}\right]e^{\mtheta^\top \vx_j}}{\sum_{j\in \mathcal{R}(T_i) } e^{\mtheta^\top \vx_j}}\right)\right]. \qedhere
\end{align*}
\end{proof}

Interestingly,  Lemma~\ref{lem: biased loss} of bias leads to an intuitive corollary regarding the unbiased estimation:
\begin{cor}\label{cor: unbiased cond}
    The mini-batch negative log-likelihood estimator is unbiased if and only if, for every sample $i\in\mathcal{B}$ with $\Delta_i=1$, the entire risk set $\mathcal{R}(T_i)$ is contained in the mini-batch $\mathcal{B}$. 
\end{cor}

\begin{proof}
According to Lemma~\ref{lem: biased loss}, the bias introduced by the mini-batch estimator is given by
$$
\mathbb{E}_\mathcal{B}\Bigl[\mathcal{L}_{\mathtt{NLL}}(\dataset,\mtheta) - \mathcal{L}_{\mathtt{NLL\text{-}mini batch}}(\dataset,\mtheta) \Bigr] 
= -\frac{1}{n}\sum_{i=1}^n \Delta_i\,\mathbb{E}_{\mathcal{B}\,|\,i\in \mathcal{B}}\!\left[\log \left(1- \frac{\sum_{j\in \mathcal{R}(T_i)} \mathbb{I}\bigl[j\notin \mathcal{B}\bigr]\,e^{\mtheta^\top \vx_j}}{\sum_{j\in \mathcal{R}(T_i)} e^{\mtheta^\top \vx_j}}\right)\right].
$$
Since the term 
$$
\frac{\sum_{j\in \mathcal{R}(T_i)} \mathbb{I}\bigl[j\notin \mathcal{B}\bigr]\,e^{\mtheta^\top \vx_j}}{\sum_{j\in \mathcal{R}(T_i)} e^{\mtheta^\top \vx_j}}
$$
is nonnegative (because \(e^{\mtheta^\top \vx_j} > 0\) and \(\mathbb{I}[j\notin \mathcal{B}] \ge 0\)), it follows that
$$
1- \frac{\sum_{j\in \mathcal{R}(T_i)} \mathbb{I}\bigl[j\notin \mathcal{B}\bigr]\,e^{\mtheta^\top \vx_j}}{\sum_{j\in \mathcal{R}(T_i)} e^{\mtheta^\top \vx_j}} \le 1.
$$
Thus, since the logarithm is an increasing function, we have
$$
\log \left(1- \frac{\sum_{j\in \mathcal{R}(T_i)} \mathbb{I}\bigl[j\notin \mathcal{B}\bigr]\,e^{\mtheta^\top \vx_j}}{\sum_{j\in \mathcal{R}(T_i)} e^{\mtheta^\top \vx_j}}\right) \le 0.
$$
In order for the expectation 
$$
\mathbb{E}_{\mathcal{B}\,|\,i\in \mathcal{B}}\!\left[\log \left(1- \frac{\sum_{j\in \mathcal{R}(T_i)} \mathbb{I}\bigl[j\notin \mathcal{B}\bigr]\,e^{\mtheta^\top \vx_j}}{\sum_{j\in \mathcal{R}(T_i)} e^{\mtheta^\top \vx_j}}\right)\right]
$$
to equal zero, the integrand (the logarithm term) must be zero almost surely for every \(i\) with \(\Delta_i = 1\) and \(i \in \mathcal{B}\).

Notice that
$$\log(x) = 0 \quad \iff \quad x = 1.$$
Thus, for each such \(i\) we require
$$1- \frac{\sum_{j\in \mathcal{R}(T_i)} \mathbb{I}\bigl[j\notin \mathcal{B}\bigr]\,e^{\mtheta^\top \vx_j}}{\sum_{j\in \mathcal{R}(T_i)} e^{\mtheta^\top \vx_j}} = 1.$$
This immediately implies
$$\frac{\sum_{j\in \mathcal{R}(T_i)} \mathbb{I}\bigl[j\notin \mathcal{B}\bigr]\,e^{\mtheta^\top \vx_j}}{\sum_{j\in \mathcal{R}(T_i)} e^{\mtheta^\top \vx_j}} = 0.$$
Since the denominator is strictly positive (as \(e^{\mtheta^\top \vx_j} > 0\) for all \(j\)), we conclude that
$$\sum_{j\in \mathcal{R}(T_i)} \mathbb{I}\bigl[j\notin \mathcal{B}\bigr]\,e^{\mtheta^\top \vx_j} = 0.$$
Given that \(e^{\mtheta^\top \vx_j}>0\) for every \(j\), the only possibility is that
$$\mathbb{I}\bigl[j\notin \mathcal{B}\bigr] = 0 \quad \text{for all } j \in \mathcal{R}(T_i),$$
or equivalently,
$$\sum_{j\in \mathcal{R}(T_i)} \mathbb{I}\bigl[j\notin \mathcal{B}\bigr] = 0.$$
This condition means that \emph{every} sample \(j\) in the risk set \(\mathcal{R}(T_i)\) must be included in the mini-batch \(\mathcal{B}\) whenever \(i\) (with \(\Delta_i=1\)) is in \(\mathcal{B}\). 
Thus, the mini-batch estimator is unbiased if and only if, for every sample \(i\) with an observed event that is in the batch, we have \(\mathcal{R}(T_i) \subseteq \mathcal{B}\). 
\end{proof}

\section{Extensions}
\label{app: extensions}
\subsection{Proportional Baseline Hazard Models -- CoxPH Variants}

The success of the CoxPH model is evident in its numerous extensions; we describe a few below. All these CoxPH-based models can be combined with the \textsc{Spectral} method as in Sec.~\ref{sec: method}.

\noindent{\bf Weighted CoxPH \cite{Reweight2015, subgroupreweight, Buchanan2014Worthreweight}:} A common extension of CoxPH is to weigh the hazard rate of each sample, yielding a loss of the form:
\begin{align}
    \NLL(\dataset,\mtheta) =  \sum_{i=1}^n {\Delta_i}\left(\log\sum_{j\in \mathcal{R}_i}\mW_{ji} e^{\mtheta^T \vx_j}- \log \mW_{ii} e^{\mtheta^T \vx_i} \right),\label{eq:weightednnl}
\end{align}
where  $\mW_{ji}\in\R_+$ denotes the weight for sample $j$ at time $T_i$. For example, \citet{Reweight2015} set weights to be a decreasing function of the size of the censored set at time $T_i$: intuitively, this weighs events less if they occur in the presence of only a few at-risk samples.  
Jakob et al. \cite{subgroupreweight} suggest using subgroup-specific weights. This allows a heterogeneous treatment against different classes: e.g., some types of cancer could be more lethal than others or different stages of cancer pose different threats to patients~\cite{hu2021subgroup, zheng2023cancergroups}. Loss~\eqref{eq:weightednnl} can again be optimized via gradient descent or second-order methods. As it already accounts for weights, Theorem~\ref{thm: ss} can be directly applied to the weighted CoxPH model. 



\noindent{\bf Heterogeneous Cox model \citep{hu2021subgroup}:} In this model, the set of samples $[n]$ is partitioned into $K$ disjoint groups $\{C_j\}_{j=1}^{(k)}$; the hazard rate then contains a group specific term:
\begin{align}
    \lambda(t|\vx_i)=\lambda_0(t)e^{(\mtheta^T \vx_i+\veta_{c_i}^T \vz_{c_i})}, \quad i\in [n], \label{hetergenuous cox}
\end{align}
where $c_i\in\{1,\ldots,K\}$ is the group $i$ belongs to, and $\veta_{c_i}$, $\vz_{c_i}$ are group-specified parameters and features, respectively. Corresponding MLE can again be performed via standard optimization methods. By replacing the linear predictor with group-specific predictors, we can again learn the parameters by applying the spectral method first to compute the intrinsic scores. 

Formally, as shown in Eq.~\eqref{hetergenuous cox}, the only difference between CoxPH and Heterogeneous Cox is there is that a part of features is associated with group-specific parameters. Thus, the solution for intrinsic scores remains the same. Thus, we modify the objective of the fitting step as
\begin{align}
    \min_{\mtheta, \veta} \sum_{i=1}^n - \vu_i^\top h_{\veta, \mtheta}(\vx_i, \vz_i)+\rho \vpi_i^{(k+1)}\log ( h_{\veta, \mtheta}(\vx_i, \vz_i)).
\end{align}
where $h_{\veta, \mtheta}(\vx_i, \vz_i) \coloneqq e^{(\mtheta^T \vx_i+\veta_{c_i}^T \vz_{c_i})}$. Therefore, we can use ADMM to alternatively update $\veta$ and $\mtheta$.


\subsection{Heterogenuous Baseline Hazard Models}
The heterogeneous baseline hazard model does not assume proportional baseline hazard functions. Therefore, the baseline hazard rate will not cancel out in the loss. Consequently, these models optimize a different likelihood. We show that by fixing baseline hazard rates as weights iteratively, we can again apply the spectral method.

\noindent{\bf Deep Heterogeneous Hazard model (DHH):}
\label{subsec: heterogenuous hazard model}
As a warmup, we present first the following simple extension to DeepSurv. We introduce heterogeneity in the dataset of the following form: we assume that samples are partitioned into $m$ classes with the sample $i$ belonging to class $c_i\in [m]$. To increase the flexibility of the standard CoxPH/DeepSurv hazard model, we consider \emph{class-specific baseline} functions. 
Formally, the DHH hazard model is defined as
\begin{align}\label{eq: DSH}
    \lambda(t|\vx_i)=\lambda_{c_i}(t) \tilde{h}_{\mtheta}( \vx_i), \quad i \in [n],
\end{align}
where $\lambda_{c_i}(\cdot)$ is the baseline hazard function of class $c_i$, and $\tilde{h}$ is again defined as in Eq.~\eqref{eq: reweight DeepSurv nll} (i.e., is either a CoxPH or \textsc{DeepSurv}  model).

Minimizing the partial likelihood under such hazard rates reduces to:
\begin{align}\label{eq: DHH problem}
    \min_{\vpi,\mtheta,\lambda} \loss_\rho(\dataset,\vpi)= &  \sum_{i=1}^n{\Delta_i} \left(   \log\sum_{j\in \mathcal{R}_i}\lambda_{c_j}(T_i) \vpi_j- \log\lambda_{c_i}(T_i) \vpi_i  \right)\\
    \textit{s.t.}\quad \vpi = & \tilde{h}_{\mtheta}(\mx)\in \R^n, \quad \vpi\geq 0,\nonumber
\end{align}



If the per-class hazard rates were known, this is precisely a weighted CoxPH/DeepSurv model:
the baseline hazard rates $\lambda_{c_j}(T_i)$ serve as fixed weights (i.e., $\mW_{ij}:=\lambda_{c_i}(T_j)\in \R, \forall i \in [m], j \in [n]$). Hence, this can directly be solved by the spectral methods we described in the previous section. However, the per-class baseline hazard rates $\lambda_{c_j}(\cdot)$ are \emph{not known}. 

Nevertheless, we can resolve this issue via alternating optimization. In particular, we can proceed iteratively, alternating between (a) estimating the class-specific baseline functions with an appropriate Breslow estimator, and then (b) estimating parameters $\mtheta$ via an application of the spectral algorithm for  Weighted CoxPH/DeepSurv. The entire process can be initialized by, e.g.,  estimating the baseline rates via Nelson-Aalen (Eq.~\eqref{eq: smoothing NA}) in the first iteration.

To summarize, at initialization, we estimate baseline rates $\lambda_{c_i}(\cdot)$ per class via Nelson-Aalen. Then, we iteratively execute:
\begin{enumerate}
    \item Given the fixed weights $\mW_{ij}:=\lambda_{c_i}(T_j)\in \R, \forall i \in [m], j \in [n]$), estimate $\mtheta$ via Alg.~\ref{algo: ADMM}.
    \item Update the baseline functions via the Breslow estimator \cite{lin2007breslow}.
    \begin{align}\label{eq:bresclass}
    \hat{\lambda}_{c_i}(t)=\frac{1}{b}\sum_{T_j\in [t-b,t+b]\cap c_j=c_i}K\left(\frac{t-T_j}{b}\right)\left(\frac{1}{\sum_{j\in \mathcal{R}_i\cap c_j=c_i}  \tilde{h}_{\mtheta}(\vx_j)}\right).
    \end{align}

    \item Repeat until convergence.
\end{enumerate}

The pseudo-code is given in Algorithm~\ref{alg: outer-loop} in \textcolor{red}{red}.


%

\noindent\textbf{Accelerated Failure Time Model (AFT) \cite{Wei1992TheAF}:}
We can apply the same alternating optimization approach to regress parameters in the  Accelerated Failure Time (AFT) model \cite{Wei1992TheAF}. In AFT, the hazard rate is given by: 
\begin{align}\label{eq: AFT}
    \lambda(t|\vx_i)=&\lambda_0(t e^{\mtheta^\top \vx_i}) e^{\mtheta^\top \vx_i}
\end{align}
where the baseline function is assumed to be accelerated by the covariates at an exponential rate. As the baseline functions depend on the samples, they will not be canceled in the partial likelihood function. Thus, the model becomes
\begin{align}
\mathcal{L}_{\mathtt{P}}(\dataset|\mtheta)&=\prod_{i=1}^n\left[\frac{\lambda(T_i|\vx_i)}{\sum_{j\in \mathcal{R}(T_i)}\lambda(T_i|\vx_j)}\right]^{\Delta_i}\\
&\stackrel{\text{Eq.}~\eqref{eq: AFT}}{=}\prod_{i=1}^n\left[   \frac{ \lambda_0(T_i e^{\mtheta^\top \vx_i}) e^{\mtheta^\top \vx_i}}{\sum_{j\in \mathcal{R}(T_i)}  \lambda_0(T_i e^{\mtheta^\top \vx_j}) e^{\mtheta^\top \vx_i}}\right]^{\Delta_j}, \label{eq: AFT plik}\\
&=\prod_{i=1}^n\left[   \frac{ \mW_{ii} e^{\mtheta^\top \vx_i}}{\sum_{j\in \mathcal{R}(T_i)}   \mW_{ji}  e^{\mtheta^\top \vx_i}}\right]^{\Delta_j},
\end{align}
where we define the weight matrix as $\mW_{ji}\coloneqq \lambda_0(T_i e^{\mtheta^\top \vx_j})$. We can again momentarilly treat this as a constant matrix, to learn $\mtheta$, and then alternate to estimate $\lambda_0$ via an appropriate Beslow estimate. 

Overall, the steps proceed as follows: In the first step, we can again first estimate $\lambda_0$ via Nelson-Aalen and set weights appropriately, by picking an initial value for parameters. Then, we iterate over:
\begin{enumerate}
    \item  Given the fixed weights $\mW_{ji}\coloneqq \lambda_0(T_i e^{\mtheta^\top \vx_j}), \forall i \in [m], j \in [n]$), estimate $\mtheta$ via Alg.~\ref{algo: ADMM}.
    \item Update the baseline functions via the Breslow estimator \cite{lin2007breslow}.
    \begin{align}\label{eq:bresAFT}
    \hat{\lambda}_{0}(t)=\frac{1}{b}\sum_{T_j\in [t-b,t+b]\cap c_j=c_i}K\left(\frac{t-T_j}{b}\right) \left(\frac{t e^{-\mtheta^\top \vx_i}}{\sum_{j\in \mathcal{R}_i}  t e^{-2\mtheta^\top x_j} }\right). 
\end{align}
    \item Repeat until convergence.
\end{enumerate}
Algorithm~\ref{alg: outer-loop} (in \textcolor{blue}{blue}) summarizes these steps.

\begin{algorithm}[!t]
\caption{Alternating Spectral Survival Analysis (\textcolor{red}{DHH}/\textcolor{blue}{AFT})}
\label{alg: outer-loop}
\begin{algorithmic}[1]
\STATE \textbf{Input:} dataset $\dataset$, penalty $\rho$, kernel $K(\cdot)$, step size $b$
\STATE Estimate baseline rates via Nelson-Aalen and initialize $\mW$
\REPEAT
    \STATE $\mtheta \gets$ SpectralWeightedSA($\vpi, \mtheta, \vu, \mW, \dataset$) \hfill \textit{// see Algorithm~\ref{algo: ADMM}}
    \STATE \textbf{Update baseline hazards:}
    \STATE \hspace{1.5em} \textcolor{red}{$\lambda_{c_i}(\cdot) \gets$ Breslow estimate (Eq.~\eqref{eq:bresclass}) $\forall i$}
    \STATE \hspace{1.5em} \textcolor{blue}{$\lambda_0(\cdot) \gets$ Breslow estimate (Eq.~\eqref{eq:bresAFT})}
    \STATE \textbf{Update weights matrix $\mW$:}
    \STATE \hspace{1.5em} \textcolor{red}{$\mW_{ji} \gets \lambda_{c_j}(T_i)$}
    \STATE \hspace{1.5em} \textcolor{blue}{$\mW_{ji} \gets \lambda_0(T_i e^{\mtheta^\top \vx_j})$}
\UNTIL{convergence}
\STATE \textbf{Output:} optimized parameters $\mtheta$
\end{algorithmic}
\noindent\textit{Note: Commands in \textcolor{red}{red} indicate pseudo-code for the \textcolor{red}{DHH} model, and commands in \textcolor{blue}{blue} indicate pseudo-code for the \textcolor{blue}{AFT} model.}
\end{algorithm}

\paragraph{Extended Hazard (EH) Model ~\citep{Ciampi1985AGM}}
\label{app:eh}
The EH model assumes that, beyond affecting hazard intensity, features also change the baseline function's time scaling. In particular, the hazard rate is
\begin{align}
\lambda(t|\vx_i)=\lambda_0(t e^{\mtheta_1^T \vx_i})e^{\mtheta_2^T \vx_i}, \quad i \in [n],\label{eq:eh}
\end{align}
 parameterized by vectors $\mtheta_1,\mtheta_2 \in \mathbb{R}^d$
The $\lambda_0(\cdot, \mtheta)$ is estimated by a kernel-smoothed Breslow estimator. 
As the baseline hazard rate now depends on the parameters, it does not vanish as it did for CoxPH in Eq.~\eqref{eq: CoxPH plik}. There are several ways to address this to train parameters $\mtheta_1$ and $\mtheta_2$ by optimizing the full likelihood \cite{EH2021}. 
The Extended Hazard model optimizes the full likelihood
\begin{align}
    &\textit{FL}(\mtheta_1, \mtheta_2, \lambda_0)\\
    =&\prod_{i=1}^n \left[ \lambda_0(T_i e^{\mtheta_1^\top \vx_i}) e^{\mtheta_2^\top \vx_i}\right]^{\Delta_i} \exp{\left[ -\Lambda_0(T_i e^{\mtheta_1^\top \vx_i}) e^{(\mtheta_2-\mtheta_1)^\top \vx_i}\right]},
\end{align}
where $\Lambda_0(t)=\int_0^t \lambda_0(s)\, ds$.

Assume the support of $\lambda_0(\cdot)$ is $[0, U]$. Then the standard way is to define $\lambda_0(\cdot)$ as a piecewise constant between $M_n$ equally partitioned intervals \cite{EH2021} (or uncensored event times \cite{lin2007breslow})
\begin{align}
    \lambda_0(t) = \sum_{j=1}^{M_n} C_j \mathbb{I}\{ \frac{(j-1)U}{M_n}\leq t\leq \frac{(j-1)U}{M_n}\}.
\end{align}

Then, after plugging the piecewise baseline function back to the full likelihood, taking the derivate w.r.t. constants to zero gives the estimates of constants $\hat{C_j}$. Then one can again obtain the optimal parameter $\hat{\mtheta}_1$ and $\hat{\mtheta}_2$, by setting the gradient w.r.t. parameters in full likelihood (with $\hat{C_j}$ plugged in) to $0$.

Eventually, the estimated parameters would contribute to a finer kernel-smoothed Breslow estimator relying on parameters for the baseline function. For more details, please refer to \cite{EH2021, lin2007breslow}.

As a special case of EH, the baseline function of AFT can be achieved by setting $\mtheta_1=0$.

\subsection{Counting Processes-based Hazard Model}
\label{app: counting}
 Survival analysis can be applied to regress events in a counting process. We present here the setting of Chen et al.~\cite{chen2023gateway}, also illustrated in Fig.~\ref{fig: counting process}. In their setting,  there are $n$ samples $\vx_1, \dots, \vx_n$, each associated with an impression made to a user (e.g., viewing a website). A dataset consists of $m$ user journeys, at which a subset of impressions are made to the user at cetrain times. Within a journey, an event of note may take place (such as, e.g., a click on an ad). The goal of regression in this setting is to predict the time until such an event happens from the history of impressions (i.e., which impressions occured and when) and their features.

 Chen et al.~\cite{chen2023gateway} model this process as follows: each impression of item $i$ is associated with  a ``clock'', i.e., a random variable with a hazard rate:
 \begin{align}\lambda(t|\vx_i)=\lambda_0 e^{h_\theta(\vx_i)}. \end{align}
 An event occurs whenever the first of these clocks expires.
Under this model, a dataset of impressions and events corresponds to the following  partial likelihood:
\begin{align}
     \NLL(\dataset,\mtheta)=-\sum_{k=1}^m \sum_{i=1}^n\Delta_{k,i} \left[ h_{\mtheta}(\vx_i)-\log \sum_{j\in \mathcal{R}_{k,i}} e^{h_{\mtheta}(\vx_j)}\right].
\end{align}
where $\Delta_{k,i}=1$ indicates an event occured in the $k$-th journey (i.e., the user clicked on the $i$-th Ad), And $\mathcal{R}_{k,i}$ denotes the ADS at risk (i.e., whose clocks have not yet expired) at the observation time of $T_{k,i}$.

This naturally maps to a sum of DeepSurv losses, on which we can apply our spectral method. 
In traditional survival analysis, each sample can at most experience the event once, and it is assumed all the samples are studied in one cohort. Thus, the optimization time complexity for partial likelihood-based methods is $\mathcal{O}(n)$, e.g., there are $n$ terms involved in the partial likelihood as in \eqref{eq: CoxPH plik}. But in the counting setting, the event (clicking ADS) may happen multiple times in different cohorts, leading to potential quadratic or exponential terms in the partial likelihood, posing scalability issues to the current survival analysis models. However, the spectral method can directly regress the intrinsic score of each Ad, reducing the complexity to linear, while outperforming the DeepSurv model.

\section{Proof of Theorem~\ref{thm: ss}}
\label{app:proof}
Denote $W_i=\{\ell|i\in R_\ell, i=\argmin_{j\in R_\ell} O_j\},L_i=\{\ell|i\in R_{\ell},i \neq \argmin_{j\in R_\ell} O_j\}$. To solve \eqref{lin pi}, we set the derivative to 0 and achieve
    \begin{align}
        \frac{\partial \loss(\vpi,\mtheta,\vu)}{\partial \vpi_i}
        =&\sum_{\ell\in W_i} {\Delta_\ell}\left(\frac{\vW_{i,i}}{\sum_{t\in \mathcal{R}_\ell}\vW_{t,\ell} \vpi_t}- \frac{1}{\vpi_i}   \right)+\sum_{\ell\in L_i} {\Delta_\ell}\left(\frac{\vW_{i,\ell}}{\sum_{t\in \mathcal{R}_\ell}\vW_{t,\ell} \vpi_t}\right) +\vu^{k}_i\\
        &+ \rho \frac{\partial D_p(\vpi||\tilde{h}_{\mtheta}(\mx))}{\partial \vpi_i}\\
        =&\sum_{\ell\in W_i} {\Delta_\ell}\left(\frac{\vW_{i,i}}{\sum_{t\in \mathcal{R}_\ell}\vW_{t,\ell} \vpi_t}- \frac{1}{\vpi_i}   \right)+\sum_{\ell\in L_i} {\Delta_\ell}\left(\frac{\vW_{i,\ell}}{\sum_{t\in \mathcal{R}_\ell}\vW_{t,\ell} \vpi_t}\right)+\sigma_i(\vpi)\label{eq: original question}\\
        =&0.\nonumber
    \end{align}
    We solve the equation above by first solving a simpler problem
    \begin{align}
        \sum_{\ell\in W_i} {\Delta_\ell}\left(\frac{\vW_{i,i}}{\sum_{t\in \mathcal{R}_\ell}\vW_{t,\ell} \vpi_t}- \frac{1}{\vpi_i}   \right)+\sum_{\ell\in L_i} {\Delta_\ell}\left(\frac{\vW_{i,\ell}}{\sum_{t\in \mathcal{R}_\ell}\vW_{t,\ell} \vpi_t}\right)=0. \label{eq: simpler eq}
    \end{align}
    By multiplying both sides with $\vpi_i$, we have
    \begin{align}
        -\sum_{\ell\in W_i} {\Delta_\ell}\left(\frac{\sum_{j\neq i\in \mathcal{R}_\ell}\vW_{j,\ell} \vpi_j}{\sum_{t\in \mathcal{R}_\ell}\vW_{t,\ell} \vpi_t}-    \right)+\sum_{\ell\in L_i} {\Delta_\ell}\left(\frac{\vW_{i,\ell} \vpi_i}{\sum_{t\in \mathcal{R}_\ell}\vW_{t,\ell} \vpi_t}\right)=0.
    \end{align}
    After rearranging the summations, we have
    \begin{align}
        &\sum_{\ell\in W_i} \sum_{j\neq i\in \mathcal{R}_\ell} {\Delta_\ell}\left(\frac{\vW_{j,\ell} \vpi_j}{\sum_{t\in \mathcal{R}_\ell}\vW_{t,\ell} \vpi_t}-    \right)=\sum_{\ell\in L_i} {\Delta_\ell}\left(\frac{\vW_{i,\ell} \vpi_i}{\sum_{t\in \mathcal{R}_\ell}\vW_{t,\ell} \vpi_t}\right)\\
        \implies & \sum_{j\neq i} \sum_{\ell \in W_i \cap L_j} {\Delta_\ell}\left(\frac{\vW_{j,\ell} \vpi_j}{\sum_{t\in \mathcal{R}_\ell}\vW_{t,\ell} \vpi_t}-    \right)=\sum_{j \neq i}\sum_{\ell\in W_j \cap L_i} {\Delta_\ell}\left(\frac{\vW_{i,\ell} \vpi_i}{\sum_{t\in \mathcal{R}_\ell}\vW_{t,\ell} \vpi_t}\right).
    \end{align}
    Therefore, denote $\mu_{ji}=\sum_{\ell\in W_i\cap L_j}\Delta_\ell \left(\frac{\vW_{j,\ell}}{\sum_{t\in R_\ell} \vW_{t,i} \vpi_t}\right)$, we have the following 
    \begin{align}
        \sum_{j\neq i} \vpi_j \mu_{ji}(\vpi)=\sum_{j \neq i} \vpi_i \mu_{ij}(\vpi),
    \end{align}
    which is the solution to \eqref{eq: simpler eq}. Moreover, according to Theorem 4.2 from \cite{yildiz20a}, we can finish the proof by connecting the solution of the simplified problem in Eq.~\eqref{eq: simpler eq} to the original problem in Eq.~\eqref{eq: original question}.\\

\section{Equivalent Max Entropy loss}
\label{app:max entropy}
While the original loss is Eq.~\eqref{eq:weighted DeepSurv lagrangian}, after removing the terms unrelated to $\mtheta$, the equivalent loss becomes
\begin{align}
    \loss_{equiv}(\vpi,\mtheta,\vu)= &  - \vu^\top \tilde{h}_{\mtheta}(\mx)+\rho \mathcal{D}_{KL}(\vpi||\tilde{h}_{\mtheta}(\mx)).
\end{align}
Therefore, after expansion over samples, the loss becomes
\begin{align}
    \loss_{equiv}(\vpi,\mtheta,\vu)
    = & \sum_{i=1}^n - \vu_i^\top \tilde{h}_{\mtheta}(\vx_i)+\rho \vpi_i^{(k+1)}\log (\tilde{h}_{\mtheta}(\vx_i))\\
    &- \rho \vpi_i^{(k+1)}\log \vpi_i^{(k+1)},
\end{align}
where the last term is irrelevant to $\mtheta$ and can be removed. Thus, we achieve the Max Entropy form of the loss.

\section{Experiment Setup}
\label{app: experiment details}
\begin{figure}[!t]
  \centering
    \includegraphics{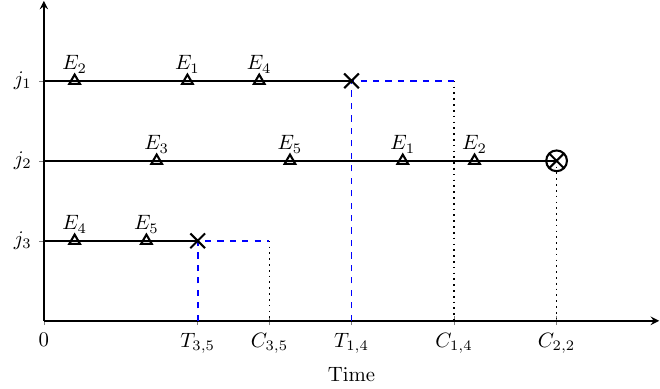}
  \caption{Illustration of the setting of Chen et al.~\cite{chen2023gateway}, modeling ADS click from survival analysis (counting process) perspective. Here $E$ denotes impressions, $\times$ denotes an event (ad click), and $j$ denotes journeys. Not all journeys lead to an ad click. The goal is to predict the time an ad is clicked from past impressions and features associated with them.}
  \label{fig: counting process}
\end{figure}


\subsection{Datasets}

Below is the summary of the datasets. There are seven vector datasets and one 3D CT scan dataset.

\noindent{\bf ADS.} As illustrated in Fig~\ref{fig: counting process}, the dataset aims at simulating the ad-clicking process. The survival time is exponentially distributed and the censoring time is uniformly distributed. The synthetic datasets consist of 100, 1k, and 10k ad journeys, where each journey may experience up to 50 ADS. The number of ADS is uniformly distributed. Meanwhile, the ADS in training, validation, and test sets have different features. The various capacities are designed to help investigate the scalability of survival analysis models.

\noindent{\bf DBCD.} The Dutch Breast Cancer Dataset (DBCD) \citep{DBCD} has a consecutive series of 295 tumors of women with breast cancer. All tumors were profiled on cDNA arrays containing 24885 genes. This initial set was reduced to a set of 4919 genes employing the Rosetta error model. Among the 295 patients, 151 were lymph node-negative (pN0) and 144 were lymph node-positive (pN+). Ten of the 151 patients who had lymph node-negative disease and 120 of the 144 who had lymph node-positive disease had received adjuvant systemic therapy consisting of chemotherapy (90 patients), hormonal therapy (20), or both (20). The median follow-up among all 295 patients was 6.7 years (range, 0.05–18.3). 

\noindent{\bf DLBCL.} The Diffuse Large B Cell Lymphoma dataset (DLBCL) \citep{DLBCL2002use} has 240 patients with 7399 genes. The gene-expression profiles of the lymphomas were used to develop a molecular predictor of survival. Subgroups with distinctive gene-expression profiles were defined on the basis of hierarchical clustering. There are three gene-expression subgroups -- germinal-center B-cell–like, activated B-cell–like, and type 3 diffuse large-B-cell lymphoma. Two common oncogenic events in diffuse large-B-cell lymphoma, bcl-2 translocation and c-rel amplification, were detected only in the germinal-center B-cell–like subgroup. Patients in this subgroup had the highest five-year survival rate. 

\noindent{\bf VDV.} The van de Vijver Microarray Breast Cancer dataset (vdv) \citep{vdv2002gene} includes 78 sporadic lymph-node-negative patients, searching for a prognostic signature in their gene expression profiles. 44 patients remained free of disease after their initial interval of at least 5 years, while 34 patients had developed distant metastases within 5 years. 4705 genes were selected from 25000 genes on the microarray.   


\noindent{\bf LUNG1.} The LUNG1 data set is publicly available at The Cancer Imaging Archive (TCIA) \citep{NLST2011national} and consists of 422 NSCLC patients. Patients with inoperable NSCLC treated at 
MAASTRO Clinic, The Netherlands, with radical radiotherapy or chemo-radiation. Here we utilize the CT images from the LUNG1 dataset with survival time information.

\noindent{\bf MovieLens.} Compiled by GroupLens Research \citep{movielens}, the MovieLens dataset is a benchmark in the movie recommender systems community. In our adaptation, user–movie interactions are enriched with temporal information to form a survival analysis setting. We treat each click on the ad as an event. The resulting dataset consists of 100k observations and we use matrix factorization on movie-user matrix to achieve 50 features. We censor 25\% of the data to simulate censoring in survival analysis study.





\subsection{Algorithm Implementation Details }
\label{app:algorithms}
{\bf \noindent Survival Analysis Regression Methods.} 
DeepSurv \citep{Katzman2018DeepSurvPT}  extends the CoxPH model by replacing the linear model with a neural network, while \textsc{FastCPH}  \citep{yang2022fastcph} further improves the implementation by introducing feature selection and considers the tie situation. \textsc{CoxCC} \citep{CoxTimeCC} enables batched gradient descent by approximating the risk set with subsets. \textsc{CoxTime} \citep{CoxTimeCC} considers time as an additional covariate and turns the relative risk function time-dependent.
DeepHit estimates the PMF while applying the ranking loss to the log-likelihood. The Neural Multi-Task Logistic Regression (N-MTLR) \citep{NMTLR} combines deep learning with MTLR, avoiding thus the use of a common baseline function. 

For all methods, except \textsc{FastCPH} , we use the DNN architectures shown in the last column Table~\ref{tab: Datasets}: this is a 6-layer 3D CNN for LUNG1 data, and MLP with 2-6 layers, of width 200, for the remaining datasets (see also App.~\ref{app:structures}); in this case, we treat the number of layers as a hyperparameter to be tuned. We also experimented with ResNet~\cite{ResNet} but did not observe a significant improvement over the MLP.

\textsc{FastCPH} has a fixed network structure, LassoNet~\cite{lassonet21}. We set the number of layers to 2-6 and the width to 200, to make this comparable to the MLP. However, its restriction to LassoNet makes it inapplicable to three-dimensional data. As such, we only apply the remaining five methods to the three-dimensional dataset LUNG1. 

All implementations of competitors are in PyTorch and obtained from the PyCox repository \cite{pycox}, except \textsc{FastCPH}  which was implemented by Yang et al. \citep{yang2022fastcph}.  Our implementation of \textsc{Spectral}, extending the code of \cite{yildiz21a}, is based on TensorFlow.

\noindent\textbf{Legacy Methods.} 
In legacy methods, imaging data are first transformed into a high-dimensional, mineable feature space (radiomics) by applying a large suite of handcrafted feature extractors. Clinical data are scalar or categorical features such as age at diagnosis, cancer stage, and histology. Haarburger et al. \cite{Haar2019} selected 8 radiomics features combined with features extracted from image patches (using CNN) and the data is then fed into a modified \textsc{DeepSurv}.  \citet{Brag2023} established multiple pipelines, while applying the 2D CNN to a 2D slide combined with radiomics and clinical data outperforms other pipelines. \citet{ZHENG2023109483} extract tumor cube from the original whole CT images and then apply 3D CNN with additional processed clinical features. \citet{zhu2016deepconvsurv} extract patches from the whole CT images based on regions of interest (ROIs) annotated with the help of pathologists and apply CNN to the patches.

\noindent{\bf \textsc{Spectral}.} As we implemented the method, we found the intrinsic scores were ill-positioned for censored samples. While it may not be apparent for other methods, the direct optimization of the partial likelihood forces the censored sample to have a 0 intrinsic score. Intuitively, as the censored sample only appears in the denominator of partial likelihood as in Eq.~\eqref{eq: CoxPH plik}, the optimal intrinsic score for the censored sample should be 0. However, it apparently should not be the case. Thus, we set $\Delta_i=1$ for all samples and treat it as regularization that keeps the intrinsic scores away from 0. Moreover, we observe that the convergence of \textsc{IterativeSpectralRanking} procedure relies on the convergence of the power method. Thus, we set the maximum iterations allowed for power method to be $50$-$200$ to ensure the convergence of the \textsc{IterativeSpectralRanking} procedure. More importantly, the power method only takes the maximum iteration in the first round, while it only requires 1-2 iteration in the following rounds. Thus, the overall running time is still comparable to the other baselines.

{\bf \noindent Hardware:} The larger models for LUNG1 are trained on Nvidia DGX with AMD EPYC 7742 64-Core Processor and A100s; and the other models are trained on internal cluster with V100-SXM2-32GB.

\noindent\textbf{Hyperparameter Search.}
For all models, we explore the learning rates ($[0.00001, \dots, 0.01, 0.1]$). For MLPs, we explore dropout rates ($[0.1,\dots, 0.5]$) and depth ($[2,\dots, 6]$). For 3D-CNN, we fix the dropout rate as $0.3$ and depth as $4$. In terms of the hyperparameters for the \textsc{Spectral}, we explore $\rho=[0.1, 0.5, 1, 2, 5, 10]$ initially and confirm the default value $\rho=1$ works better generally as in Sec.~\ref{sec: experiments}. Moreover, we explore the maximum iteration of the power method from $[10,20,50, 100,200]$. 

We use full-batch GD for all one-dimension datasets, where all the samples are computed for gradient descent, computing the full-loss (Eq.~\eqref{eq: CoxPH plik}) for competitors. For CT dataset LUNG1, as the data volume increases exponentially in the dimension, the GPU cannot handle all the samples simultaneously. Therefore, we compute the loss Eq.~\eqref{eq: CoxPH plik} in mini-batches, which includes $10\%$ of the dataset in each batch; all competitors crash (see also Table~\ref{tab: LUNG1 performance.}) when this batch is increased. No batching whatsoever is needed for Eq.~\eqref{lin pi} (that involves the data) for \textsc{Spectral}; we use a fixed batch size of $16$ for the NN fit in Eq.~\eqref{lin theta} (that is data independent).

Optimal hyperparameters are selected based $k$-fold cross-validation on the realistic vector dataset, using also early stopping; we use CI as a performance metric. For LUNG1 and Synth ADS, we have fixed split train, validation, and test sets. For LUNG1, it is 70\%, 15\%, and 15\%. For Synth ADS, due to the large amount of data, we have fixed the validation and test sets to be 1000 journeys. As we apply early stopping, we evaluate the method with the checkpoint that shows the best validation performance (CI).
\subsection{Metrics}
\label{app:metrics}

In terms of the predictive performance, we adopt the three metrics, CI, AUC, and RMSE.

We evaluate the models with the predicted survival function $\hat{S}(\cdot)$ according to  $$S(t|\vx)=\exp{\left(-\int_0^t \lambda(s|\vx)\, ds\right)}.$$ and \eqref{eq: smoothing NA} with respect to the following metrics.

\noindent{\bf The concordance index (C-index).} C-index is one of the most common evaluation metrics in survival analysis, which estimates the concordant probability over all the observation pairs.
\begin{equation}
    \text{C-index} = P\left[ \hat{S}(T_i|\vx_i)< \hat{S}(T_i|\vx_j)|T_i<T_j, \Delta_i=1 \right]
\end{equation}


\noindent{\bf Cumulative AUC.} The Cumulative AUC is achieved by integrating the AUC value at specific time points as in IBS. The AUC at time $t$ is defined as
\begin{equation}
    AUC(t) =
\frac{\sum_{i=1}^n \sum_{j=1}^n \ind(y_j > t) \ind(y_i \leq t) \omega_i
\ind(\hat{h}(\mathbf{x}_j) \leq \hat{h}(\mathbf{x}_i))}
{(\sum_{i=1}^n I(y_i > t)) (\sum_{i=1}^n \ind(y_i \leq t) \omega_i)},
\end{equation}
where $\hat{h}(\vx_i)$ is the estimated relative risk score of $\vx_i$ and $\omega_i=1/\hat{G}(y_i)$ is the IPCW.
Similarly, we integrate the time-dependent AUC over the test set time interval $t\in[tmin,tmax]t\in [t_{\min}, t_{\max}]$,
\begin{equation}
    iAUC = \frac{1}{t_{\max}-t_{\min}}\int_{t_{\min}}^{t_{\max}} \mathrm{AUC}(t) d t,
\end{equation}
where we refer to it as iAUC (integrated AUC) due to its integral property.

\noindent{\bf RMSE.} We follow the approach of \citet{SurvCNN21} by estimating the ground-truth survival function on the test set using the Kaplan-Meier estimator $S_{KM}(t)$, defined as:
$S_{KM}(t) = \prod_{t_j \leq t} \left(1-\frac{d_j}{n_j}\right),$
where $d_j$ is the number of events at time $t_j$, and $ n_j$ is the number of samples at risk just prior.
Then, the Root Mean Square Error (RMSE) for survival functions is given by:

\begin{equation}
\text{RMSE} = \sqrt{\frac{1}{T} \sum_{t=1}^{T} \left( S_{KM}(t) - \hat{S}(t) \right)^2 }
\end{equation}

where $\hat{S}(t)$ is the predicted survival function, while $t\in {1,\dots,T}\in [t_{\min}, t_{\max}]$ is evenly distributed.

\noindent{\bf Scalability metrics.} We measure the memory and per sample run-time.

We measure the peak usage of the memory in MB for each method, including the computation of neural networks, losses, and \textsc{Spectral} methods.

As for the per sample run-time, we record the time of training and evaluating altogether and report the total time in seconds.

\subsection{Network Structures}
\label{app:structures}
Both networks are optimized through Adam optimizer \cite{adam}. We provide the structure summary below.
\paragraph{LUNG1: 3D-CNN}
For the LUNG1 dataset, we implemented a 6-layer 3D-CNN that consists of 4 Conv FC layers and 2 dense layers. The details are shown in Table~\ref{tab: 3DCNN}.
\begin{table}[h]
\centering
\begin{tabular}{@{}ll@{}}
\toprule
Layer & Configuration \\
\midrule
Input & $64 \times 128 \times 128 $ CT image \\
\midrule
Block 1 & \begin{tabular}[c]{@{}l@{}}Conv3D: 64 filters, $3 \times 3 \times 3$ kernel, ReLU activation\\ Max pooling: $2 \times 2 \times 2$ \\ Batch normalization\end{tabular} \\
\midrule
Block 2 & \begin{tabular}[c]{@{}l@{}}Conv3D: 64 filters, $3 \times 3 \times 3$ kernel, ReLU activation\\ Max pooling: $2 \times 2 \times 2$ \\ Batch normalization\end{tabular} \\
\midrule
Block 3 & \begin{tabular}[c]{@{}l@{}}Conv3D: 128 filters, $3 \times 3 \times 3$ kernel, ReLU activation\\ Max pooling: $2 \times 2 \times 2$ \\ Batch normalization\end{tabular} \\
\midrule
Block 4 & \begin{tabular}[c]{@{}l@{}}Conv3D: 256 filters, $3 \times 3 \times 3$ kernel, ReLU activation\\ Max pooling: $2 \times 2 \times 2$ \\ Batch normalization\end{tabular} \\
\midrule
GlobalAvgPool & Global average pooling \\
Dense & Dense layer with 512 neurons, ReLU activation \\
Dropout & Dropout with rate 0.3 \\
Output & Dense layer with 1 neuron, sigmoid activation \\
\bottomrule
\end{tabular}
\caption{Overview of the structure of 3D-CNN for LUNG1 dataset}
\label{tab: 3DCNN}
\end{table}

\paragraph{Vector datasets}
For vector datasets, we implemented an MLP with 2 to 6 layers. The details are shown in Table~\ref{tab: MLP}.
\begin{table}[h]
\centering
\begin{tabular}{@{}ll@{}}
\toprule
Layer & Configuration \\
\midrule
Input & $d$ dimension features \\
\midrule
Input layer & \begin{tabular}[c]{@{}l@{}}Dense: $200$ neurons, ReLU activation,\end{tabular} \\
 & Dropout with rate $dropout$ \\
\midrule
\multicolumn{2}{l}{\textbf{Repeat for $depth - 1$ times:}} \\
Hidden layers & \begin{tabular}[c]{@{}l@{}}Dense: $200$ neurons, ReLU activation, \end{tabular} \\
 & Dropout with rate $dropout$ \\
\midrule
Output & \begin{tabular}[c]{@{}l@{}}Dense: 1 neuron, Sigmoid activation, \end{tabular} \\
\bottomrule
\end{tabular}
\caption{Overview of the structure of 1D Multilayer Perceptron (MLP)}
\label{tab: MLP}
\end{table}

\section{Scalability and Sensitivity Experiments}\label{app: additional experiments}
\subsection{Scalability regarding $n$ and $d$}
In addition to the scalability analysis in Sec.~\ref{sec: experiments}, we evaluate runtime and memory scalability of the proposed method across a diverse set of datasets with varying numbers of features ($d$) and samples ($n$). As shown in Table~\ref{tab: runtime and memory}, \textsc{Spectral} scales efficiently in both $n$ and $d$, achieving consistently lower GPU memory usage and runtime, especially in high-dimensional (e.g., LUNG1, DBCD) and large-sample (e.g., ADS10k, MovieLens) settings where other methods fail or do not scale as well.

\begin{table}[t]
\caption{Comparison of \textsc{Spectral} with other state-of-the-art (SOTA) hazard models on 8 datasets in terms of runtime and memory. ``--'' indicates that the method is not applicable to the dataset and \ding{56} indicates an out-of-memory failure.}
\label{tab: runtime and memory}
\begin{center}
\begin{small}
\resizebox{\textwidth}{!}{
\begin{tabular}{cccccccccc}
\toprule
 &  & \textbf{DBCD} &  \textbf{DLBCL} & \textbf{VDV} & \textbf{LUNG1} & \textbf{ADS100} & \textbf{ADS1k} & \textbf{ADS10k} & \textbf{MovieLens} \\
\midrule
\textbf{\# Features ($d$)} &  & 4919 & 7399 & 4705 & 17M & 50 & 50 & 50 & 200 \\
\textbf{\# Samples ($n$)} &  & 295 & 240 & 78 & 422 & 100 & 1K & 10K & 100K \\
\midrule
\multirow{7}{*}{\textbf{Memory (MB)} $\downarrow$}
& \textsc{DeepHit} \cite{lee2018deephit}     & 489    & 507    & 487    & 61GB(20\%BS) & --    & --     & --      & -- \\
& \textsc{DeepSurv} \cite{Katzman2018DeepSurvPT} & 477    & 539    & 571    & 61GB(20\%BS) & 429   & 565    & 2844    & 1611 \\
& \textsc{FastCPH} \cite{yang2022fastcph}      & 589    & 683    & 569    & --           & --    & --     & --      & -- \\
& \textsc{CoxTime} \cite{CoxTimeCC}            & 591    & 687    & 573    & \ding{56}   & --    & --     & --      & -- \\
& CoxCC \cite{CoxTimeCC}                       & 549    & 619    & 529    & \ding{56}   & --    & --     & --      & -- \\
& \textsc{NMTLR} \cite{NMTLR}                   & 529    & 609    & 507    & 61GB(20\%BS) & --    & --     & --      & -- \\
& \textsc{Spectral}                           & 451    & 473    & 431    & 15GB(100\%BS)& 429   & 473    & 618     & 1424 \\
\midrule
\multirow{7}{*}{\textbf{Runtime (s)} $\downarrow$} 
& \textsc{DeepHit} \cite{lee2018deephit}     & 64.19 & 85.11 & 62.01 & 211.22 & --    & --     & --      & -- \\
& \textsc{DeepSurv} \cite{Katzman2018DeepSurvPT} & 120.34 & 171.86 & 99.57 & 2038   & 35.7  & 197.4  & 3029.91 & 240 \\
& \textsc{FastCPH} \cite{yang2022fastcph}      & 257.10 & 290.87 & 258.33 & --     & --    & --     & --      & -- \\
& \textsc{CoxTime} \cite{CoxTimeCC}            & 58.85  & 89.53  & 99.57  & \ding{56} & --    & --     & --      & -- \\
& CoxCC \cite{CoxTimeCC}                       & 61.63  & 84.99  & 55.35  & \ding{56} & --    & --     & --      & -- \\
& \textsc{NMTLR} \cite{NMTLR}                   & 60.86  & 73.63  & 55.40  & 256.52 & --    & --     & --      & -- \\
& \textsc{Spectral}                           & 65.17  & 90.09  & 63.54  & 760.45 & 15.77 & 62.38  & 199.5   & 172 \\
\bottomrule
\end{tabular}
}
\end{small}
\end{center}
\end{table}
}{}



\end{document}